\documentclass[11pt,a4paper]{article}

\usepackage{arxiv/insai-preprint}
\graphicspath{{arxiv/}}
\usepackage{amsmath,amssymb}
\usepackage{booktabs}
\usepackage{colortbl}
\usepackage{longtable}
\usepackage{multirow}
\usepackage{array}
\usepackage{tabularx}
\usepackage{pifont}
\usepackage{placeins}
\usepackage{xurl}
\newcolumntype{Y}{>{\centering\arraybackslash}X}
\definecolor{rankBest}{HTML}{E8BF9F}
\definecolor{rankSecond}{HTML}{F7E6D8}
\DeclareRobustCommand{\rankbest}[1]{\cellcolor{rankBest}\textbf{#1}}
\DeclareRobustCommand{\ranksecond}[1]{\cellcolor{rankSecond}\underline{#1}}

\usepackage[numbers,sort&compress]{natbib}
\usepackage{chapterbib}
\usepackage[pagebackref,breaklinks,colorlinks,hypertexnames=false,
  allcolors=ReportAccent,pdfversion=1.7]{hyperref}

\renewcommand{\ReportShortTitle}{GOLD-SLAM}
\newcommand{\PaperSubtitle}{Gaussian-splatting Omnidirectional LiDAR-Depth SLAM}
\title{\ReportShortTitle: \PaperSubtitle}
\hypersetup{
  pdftitle={\ReportShortTitle: \PaperSubtitle},
  pdfauthor={Weicong Zhang, Yuan Huang, Yu Zheng, Lin Chenxiao, Yingying Li,
    Lilong Fan, Liangdong Zhuang, Yufan Chen, Ruiping Liu, Junwei Zheng,
    Jiaming Zhang}
}

\newcommand{\ResourceLink}[3]{%
  \raisebox{-1.4pt}{\includegraphics[width=9pt,height=9pt,keepaspectratio]{fig/#1.pdf}}%
  & \textbf{#2} & \href{#3}{\nolinkurl{#3}}%
}

\renewcommand{\ReportResources}{%
  \smallskip
  {\color{ReportRule}\hrule height 0.5pt}%
  \vspace{5pt}%
  {\sffamily\footnotesize
    \noindent\begin{tabular}{@{}c@{\hspace{5pt}}l@{\hspace{9pt}}l@{}}
      \ResourceLink{brand-github}{GitHub}{https://github.com/InSAI-Lab/GOLD-SLAM}\\
      \ResourceLink{brand-hugging-face}{Hugging Face}{https://huggingface.co/datasets/InSAI-Lab/WHOLE}\\
      \ResourceLink{gold_slam_icon}{Project page}{https://insailab.org/GOLD-SLAM}\\
      \ResourceLink{brand-insai}{InSAI Lab}{https://insailab.org/}
    \end{tabular}\par
    \smallskip
    \noindent\textbf{Date}\enspace November 2026\par}%
}

\newcommand{\GoldAffiliation}[2]{\mbox{\textsuperscript{#1}\,#2}}

\newcommand{\GoldInstitutionLogo}[2]{%
  \raisebox{-0.5\height}{\includegraphics[height=#1]{fig/brand-#2.pdf}}%
}

\newcommand{\GoldAuthor}[2]{\mbox{\textbf{#1}\textsuperscript{#2}}}

\newcommand{\GoldTitle}{%
  \ReportTitleBlock{%
    \raisebox{-7.6pt}{\includegraphics[height=40.25pt]{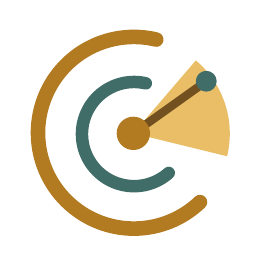}}%
    \hspace{4.3pt}\ReportShortTitle:%
  }{\PaperSubtitle}{%
    \GoldAuthor{Weicong Zhang}{1,2}\enspace
    \GoldAuthor{Yuan Huang}{1}\enspace
    \GoldAuthor{Yu Zheng}{1}\enspace
    \GoldAuthor{Lin Chenxiao}{1}\enspace
    \GoldAuthor{Yingying Li}{2}\enspace
    \GoldAuthor{Lilong Fan}{3}\\[1pt]
    \GoldAuthor{Liangdong Zhuang}{2,\textdagger}\enspace
    \GoldAuthor{Yufan Chen}{4,5}\enspace
    \GoldAuthor{Ruiping Liu}{4}\enspace
    \GoldAuthor{Junwei Zheng}{4}\enspace
    \GoldAuthor{Jiaming Zhang}{1,\textdagger}\par
    \vspace{3pt}%
    {\footnotesize\bfseries
      \noindent
      \GoldAffiliation{1}{Hunan University}\quad
      \GoldAffiliation{2}{Tsinghua University}\\
      \GoldAffiliation{3}{China Railway Construction Bridge Engineering Bureau Group Co., Ltd.}\\
      \GoldAffiliation{4}{Karlsruhe Institute of Technology}\quad
      \GoldAffiliation{5}{University of Oxford}\hfill
      {\mdseries\color{ReportMuted}\textsuperscript{\textdagger} Corresponding authors}\par}%
    \vspace{5pt}%
    \noindent
    \hspace*{4pt}%
    \GoldInstitutionLogo{28pt}{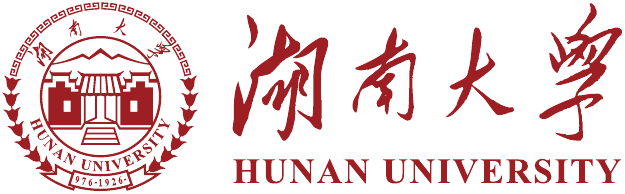}\hfill
    \GoldInstitutionLogo{27pt}{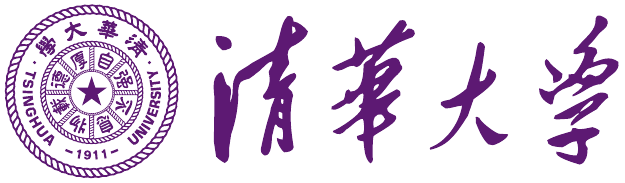}\hfill
    \GoldInstitutionLogo{34pt}{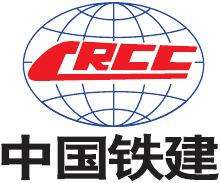}\hfill
    \GoldInstitutionLogo{24pt}{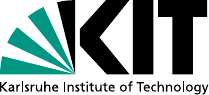}\hfill
    \GoldInstitutionLogo{30pt}{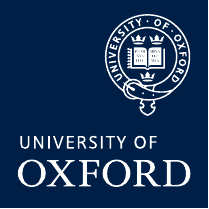}%
    \hspace*{4pt}\par
  }%
}

\begin{document}
\begingroup
\GoldTitle
\begin{abstract}
Panoramic cameras capture nearly the entire spherical field of view in a single frame.
However, the mismatch between this panoramic coverage and the field of view of LiDAR poses a substantial challenge for panoramic Gaussian splatting SLAM that relies on LiDAR depth priors.
We introduce \textbf{GOLD-SLAM}, a \underline{G}aussian-splatting \underline{O}mnidirectional \underline{L}iDAR-\underline{D}epth SLAM system that combines high-quality rendering with accurate geometric reconstruction.
It performs tracking and mapping in the spherical domain and uses tracked visual features in regions without LiDAR coverage to improve pose estimation.
To address insufficient coverage of newly observed surfaces and floating artifacts during incremental mapping, we introduce Depth-Aware Map Extension and Pruning (DAEP).
To enable broader evaluation, we collect a new dataset, \textbf{WHOLE}, using a UAV platform across \underline{W}eak-geometry, \underline{H}igh-dynamic, \underline{O}mnidirectional, \underline{L}ow-light, and \underline{E}asy scenarios.
Extensive experiments on public benchmarks and WHOLE demonstrate competitive rendering quality and improved geometric accuracy relative to perspective systems.
Compared with existing panoramic baselines, GOLD-SLAM improves mean PSNR by 5.56--9.10\,dB and reduces mean Depth-L1 by 71.9--83.9\%.
To assess deployment feasibility on resource-constrained platforms such as UAVs, we evaluate an embedded configuration of GOLD-SLAM on a Jetson Orin NX (16\,GB), demonstrating online operation at the cost of reconstruction quality.
\end{abstract}

\section{Introduction}
\label{sec:intro}

Ground and aerial robots require reliable localization, dense geometry, and photorealistic rendering for interaction, often under a limited sensor budget~\cite{kinectfusion,imap}.
LiDAR provides sparse geometric measurements with metric scale, IMUs supply high-frequency inertial constraints, and cameras capture dense color and texture.
Their complementary roles underpin multisensor odometry and mapping systems~\cite{licfusion,lvisam,r2live,fastlivo,fastlivo2}.
3D Gaussian Splatting (3DGS)~\cite{kerbl3dgs} represents scenes using explicit anisotropic Gaussians.
Its efficient rasterization enables faster rendering than NeRF~\cite{mildenhall2020nerf},
promoting the development of Gaussian-based SLAM~\cite{monogs,photoslam,gaussianlic,gaussianlic2,odgsslam}.

However, the narrow field of view (FoV) of perspective cameras limits scene coverage and inter-frame correspondences, making pose estimation more vulnerable to repetitive or sparse textures and rapid motion~\cite{panodirectvo}.
Panoramic cameras expand coverage to a $360^\circ\times180^\circ$ spherical FoV, but their projections introduce stronger nonlinear distortion~\cite{360vo}.
ODGS-SLAM~\cite{odgsslam} extends incremental Gaussian SLAM to equirectangular projection (ERP) images.
However, it suffers from scale drift outdoors, motivating the use of LiDAR geometric constraints for large-scale mapping.

\begin{figure}[!htbp]
  \centering
  \includegraphics[width=\linewidth]{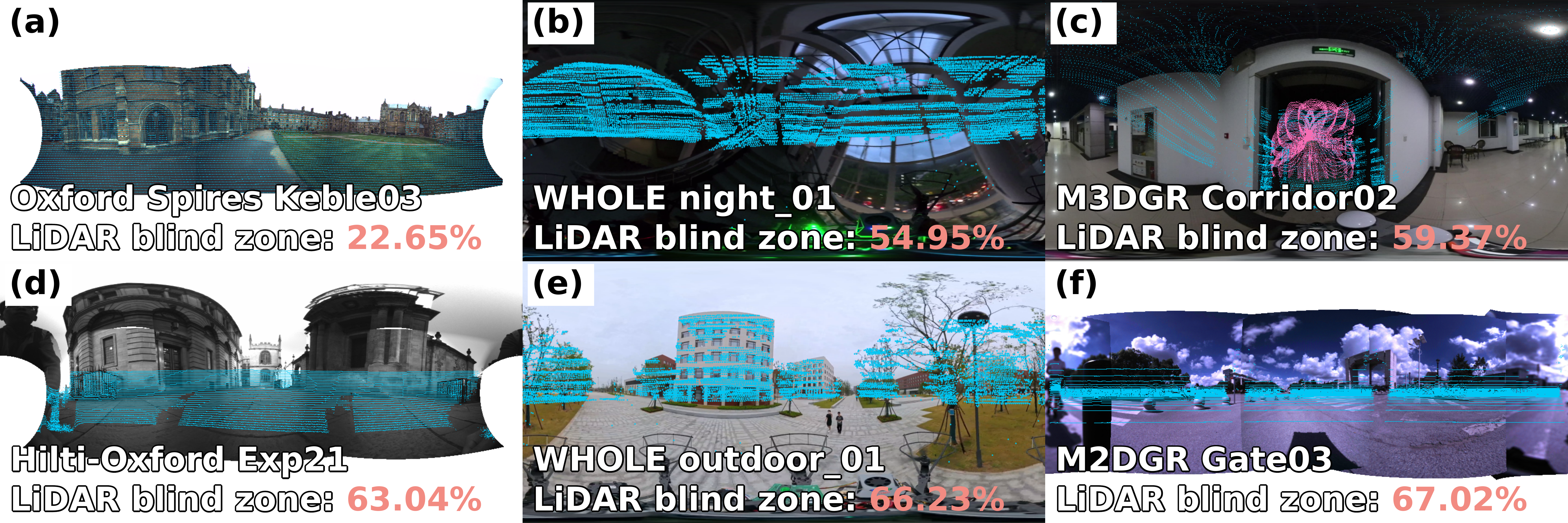}
  \caption{The limited LiDAR coverage (colored points) in 360{\textdegree} panoramic images
  leads to large LiDAR blind-zone (percentages are solid-angle weighted) and multimodal fusion challenge.}
  \label{fig:erp-valid-domain}
\end{figure}

\Needspace{11\baselineskip}
Panoramic LiDAR--inertial--visual Gaussian SLAM faces several technical challenges:
\begin{enumerate}
  \item \textbf{Nonuniform ERP sampling and polar distortion.} ERP projection introduces nonuniform sampling on the sphere and substantial stretching near the poles~\cite{odgsslam,panogs_slam}.
  \item \textbf{Greater impact of the LiDAR blind zone.} LiDAR--camera FoV mismatch leaves regions without depth support~\cite{gaussianlic2,hdpvslam}.
  Panoramic coverage exposes more such regions, complicating Gaussian initialization (Fig.~\ref{fig:erp-valid-domain}).
  \item \textbf{Interference with new surface reconstruction.} In incremental Gaussian SLAM, high accumulated opacity from existing Gaussians can block opacity-only insertion on newly observed nearby surfaces~\cite{splatam,gsslam}.
\end{enumerate}

Beyond these technical challenges, commonly used LiDAR--visual benchmarks focus on ground-robot or handheld/backpack acquisition~\cite{m2dgr,m3dgr,hiltioxford,oxfordspires}, leaving UAV motion underrepresented.
In this context, our contributions are:
\begin{itemize}
  \item We present GOLD-SLAM, a Gaussian splatting SLAM framework fusing LiDAR, IMU, and panoramic cameras with a shared spherical model for tracking and mapping.
  \item We combine blind-zone visual constraints with Patch Brightness Uncertainty (PBU) to improve pose estimation under limited LiDAR coverage and illumination changes.
  \item We combine measured LiDAR depth with panoramic depth predictions for Gaussian initialization and design Depth-Aware Map Extension and Pruning (DAEP) to improve incremental surface reconstruction.
  \item We collect WHOLE, a dataset of LiDAR and IMU measurements and panoramic images acquired with a customized UAV platform in indoor, outdoor, and degraded environments.
  Results on WHOLE and public benchmarks demonstrate competitive localization and improved geometric reconstruction with broad scene coverage.
\end{itemize}

\begin{figure}[!htbp]
  \centering
  \includegraphics[width=\textwidth]{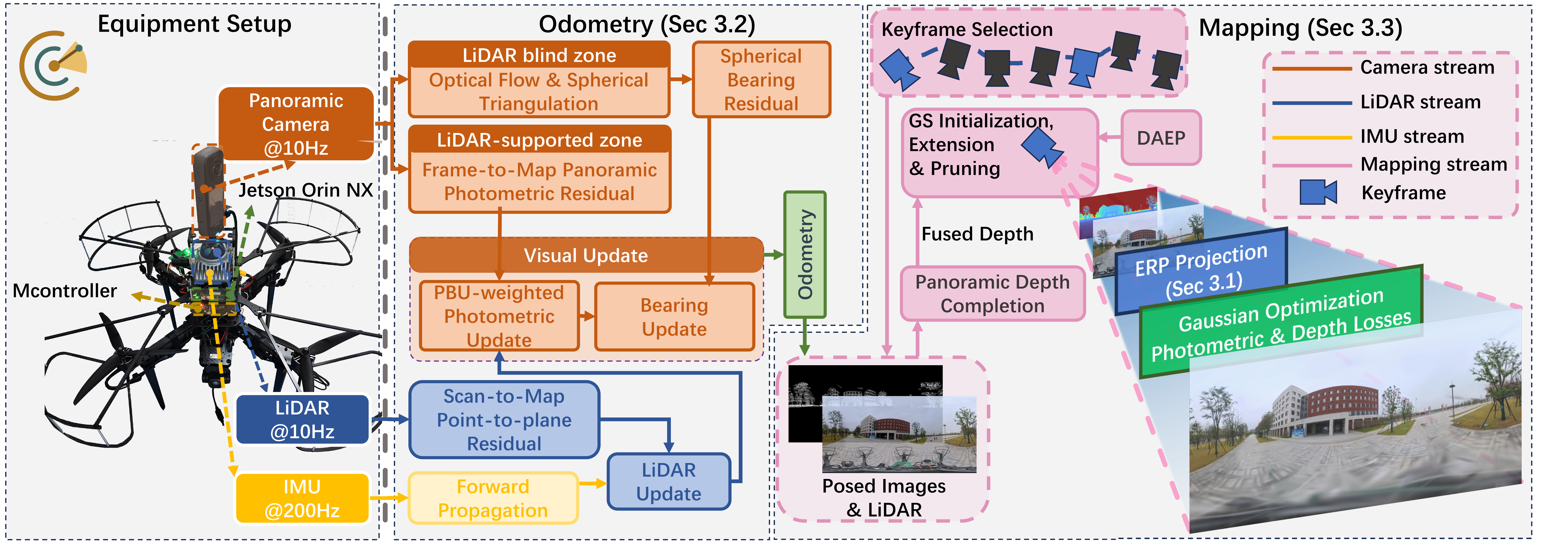}
  \caption{\textbf{Overview of GOLD-SLAM}. The UAV platform collects LiDAR, IMU measurements and panoramic images.
  Inspired by FAST-LIVO2's error-state filtering framework~\cite{fastlivo2}, the frontend integrates panoramic visual updates with IMU propagation and LiDAR scan-to-map constraints.
  Inter-frame constraints use features in the LiDAR blind zone, while Patch Brightness Uncertainty (PBU) models shared brightness variations in the LiDAR-supported zone.
  To support panoramic reconstruction, we adapt the Gaussian-LIC2 backend~\cite{gaussianlic2} to spherical projection and rendering.
  Gaussians are initialized using measured LiDAR depth and predicted panoramic depth, and optimized with photometric and LiDAR depth losses.
  Depth-Aware Map Extension and Pruning (DAEP) combines depth-aware insertion, depth-guided adaptive updates, and depth-conflict pruning to improve the reconstruction of newly observed surfaces.}
  \label{fig:system-overview}
\end{figure}

\section{Related Work}
\label{sec:related}

\textbf{Gaussian Splatting SLAM.}
3DGS~\cite{kerbl3dgs} enables efficient photorealistic rendering through differentiable Gaussian rasterization.
MonoGS~\cite{monogs} uses differentiable splatting for monocular SLAM, with analytic pose Jacobians and optional depth supervision.
Photo-SLAM~\cite{photoslam} combines geometric and photometric features with progressive Gaussian pyramid learning and geometric densification for monocular, stereo, and RGB-D SLAM.
SplaTAM~\cite{splatam} and GS-SLAM~\cite{gsslam} use depth-guided Gaussian insertion, with GS-SLAM also suppressing floating artifacts.
Unlike these perspective systems, GOLD-SLAM performs panoramic reconstruction with LiDAR geometric supervision and predicted depth for blind-zone Gaussian initialization.

\noindent\textbf{Omnidirectional SLAM.}
Omnidirectional SLAM expands visual coverage through multi-camera systems or fisheye cameras.
Gaussian projection and rendering must handle polar stretching and horizontal periodicity in ERP images.
360ORB-SLAM~\cite{orb360slam} combines panoramic features, triangulation, and depth completion for localization and point-cloud mapping.
Panoramic LiDAR DSO~\cite{panodirectvo} associates panoramic images with LiDAR observations for robust tracking.
ODGS~\cite{odgs} reconstructs Gaussian scenes from ERP images, and ODGS-SLAM~\cite{odgsslam} extends incremental Gaussian SLAM to ERP input.

\noindent\textbf{Multimodal Fusion.}
In large-scale scenes, multimodal fusion supports localization and reconstruction under visual degradation or LiDAR degeneracy.
LIV-GaussMap~\cite{livgaussmap} uses LiDAR--IMU poses and initial geometry, then refines Gaussians photometrically.
Gaussian-LIC~\cite{gaussianlic} combines tightly coupled LiDAR--IMU--camera localization with Gaussian initialization from LiDAR points and blind-zone visual triangulation.
GS-LIVM~\cite{gslivm} completes LiDAR surfaces via Gaussian process regression, using predicted covariance for Gaussian initialization.
Gaussian-LIC2~\cite{gaussianlic2} combines lightweight depth completion for blind-zone Gaussian initialization with LiDAR depth supervision.
Building on this, GOLD-SLAM supports panoramic reconstruction and uses blind-zone visual constraints and PBU for pose estimation.

\section{Method}
\label{sec:method}

GOLD-SLAM uses shared spherical geometry to fuse LiDAR, IMU, and panoramic vision for pose estimation and guide Gaussian mapping with LiDAR depth (Fig.~\ref{fig:system-overview}).

\begin{figure}[!htbp]
  \centering
  \includegraphics[width=0.73\linewidth]{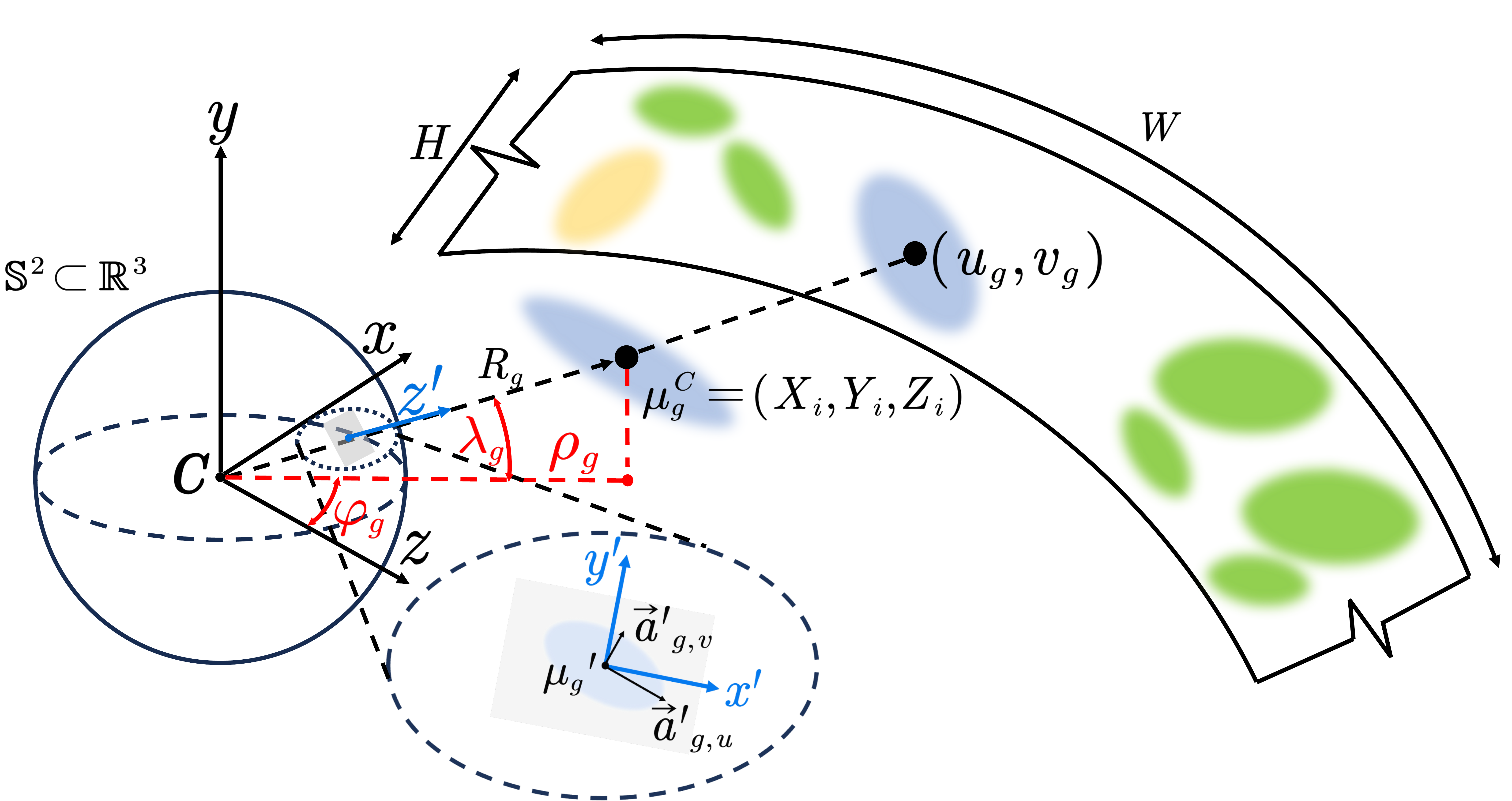}
  \caption{\textbf{The unified spherical projection geometry}.}
  \label{fig:spherical-projection}
\end{figure}

\subsection{Unified Spherical Geometry}
\label{sec:erp-camera}

The spherical projection geometry is illustrated in Fig.~\ref{fig:spherical-projection}.
Superscripts $W$, $I$, and $C$ denote the world, IMU, and camera frames, respectively, and $\mathbf R_A^B$ denotes the rotation from frame $A$ to frame $B$.
For a point $\boldsymbol\mu^C=(x,y,z)^\mathsf T$, we define radial distance $R=\|\boldsymbol\mu^C\|_2$, longitude $\varphi=\operatorname{atan2}(x,z)$, and latitude $\lambda=\operatorname{atan2}(y,\rho)$, where $\rho=\sqrt{x^2+z^2}$.
For a $W\times H$ ERP image, projection and back-projection are
\begin{equation}
  \pi(\boldsymbol\mu^C)=
  \begin{bmatrix}
    u\\
    v
  \end{bmatrix}
  =
  \begin{bmatrix}
  \frac{W}{2\pi}(\varphi+\pi)\\
  H(\frac{1}{2}-\frac{\lambda}{\pi})
  \end{bmatrix},\;
  \boldsymbol\mu^C=R\mathbf d(u,v),
  \label{eq:erp-projection}
\end{equation}
where $\mathbf d(u,v)=[\cos\lambda\sin\varphi,\sin\lambda,\cos\lambda\cos\varphi]^\mathsf T$ is the unit bearing vector in the camera frame.
For back-projection, $\varphi=2\pi u/W-\pi$ and $\lambda=\pi(1/2-v/H)$.

The Jacobian of Eq.~\eqref{eq:erp-projection}, $\mathbf J_\pi=\partial\pi/\partial\boldsymbol\mu^C$~\cite{odgsslam}, is used for panoramic pose updates and Gaussian covariance projection.
Coordinate transformations and perturbation conventions are given in Appendix~\ref*{sec:supp-state-pose}.

For Gaussian rendering, the map $\mathcal G$ contains $N_{\rm GS}$ primitives, each with a world-frame center $\boldsymbol\mu_g$, covariance $\boldsymbol\Sigma_g$, opacity $o_g$, and spherical harmonics (SH) coefficients $\mathbf f_g$.
Under the local tangent-plane approximation~\cite{odgs,odgsslam}, the camera-frame covariance $\boldsymbol\Sigma_g^C=\mathbf R_W^C\boldsymbol\Sigma_g(\mathbf R_W^C)^\mathsf T$ is projected to ERP image space as
\begin{equation}
  \boldsymbol\Sigma^{\mathrm{img}}_g
  =\mathbf J_{\pi,g}\boldsymbol\Sigma_g^C\mathbf J_{\pi,g}^\mathsf T+\epsilon\mathbf I,
  \label{eq:covariance-projection}
\end{equation}
where $\mathbf J_{\pi,g}$ is evaluated at the Gaussian center and the low-pass term $\epsilon>0$ prevents excessively small projections~\cite{kerbl3dgs}.

\subsection{Panoramic LiDAR--Inertial--Visual Odometry}
\label{sec:panoramic-odometry}

The LiDAR frontend uses voxel-based plane maps~\cite{voxelmap}, selecting 0.5, 1, or 2 m base voxel sizes online (Appendix~\ref*{sec:supp-adaptive-voxel}).

\textbf{Visual Constraints in the LiDAR Blind Zone.}
We extract corners in the LiDAR blind zone and track them across frames using pyramidal optical flow.
Forward--backward flow consistency rejects incorrect correspondences.
With sufficient parallax, we triangulate each feature $i$ from historical bearing rays (Appendix~\ref*{sec:supp-triangulation}).
Transforming the triangulated point $\mathbf P_i^W$ into the current camera frame $k$ gives $\mathbf p_{ik}^C$.
The predicted and measured bearings are $\widehat{\mathbf d}_{ik}=\mathbf p_{ik}^C/\|\mathbf p_{ik}^C\|_2$ and $\mathbf d_{ik}=\mathbf d(u_{ik},v_{ik})$, respectively.
We project the bearing error onto the tangent plane at $\mathbf d_{ik}$:
\begin{equation}
  \mathbf r_{ik}=\mathbf B_{ik}(\widehat{\mathbf d}_{ik}-\mathbf d_{ik}),
  \label{eq:blind-bearing-residual}
\end{equation}
where $\mathbf B_{ik}$ has orthonormal rows spanning this plane.

\textbf{Uncertainty-Weighted Fusion.}
We estimate each triangulated point's covariance from parallax and historical pose uncertainty, retaining a shared pose-uncertainty component across views.
Propagating this covariance through Eq.~\eqref{eq:blind-bearing-residual} and adding angular observation noise yields the bearing-residual covariance $\boldsymbol\Sigma_{r,i}\in\mathbb R^{2\times2}$.
The residuals contribute an information matrix $\boldsymbol\Lambda_V=\sum_i\mathbf H_i^\mathsf T\boldsymbol\Sigma_{r,i}^{-1}\mathbf H_i$, where $\mathbf H_i$ is the Jacobian of the residual $\mathbf r_{ik}$ with respect to the error state.
Let $\mathbf P_k=\mathbf L\mathbf L^\mathsf T$ be the state covariance after the LiDAR and direct visual updates, with $\mathbf L$ its Cholesky factor.
To limit the relative visual information, we scale both $\boldsymbol\Lambda_V$ and the corresponding weighted residual term by $\eta$:
\begin{equation}
  \eta=\min\left(1.4,\frac{\gamma}
  {\lambda_{\max}(\mathbf L^\mathsf T\boldsymbol\Lambda_V\mathbf L)}\right),
  \label{eq:blind-information-cap}
\end{equation}
where $\lambda_{\max}$ denotes the largest eigenvalue and $\gamma$ bounds the relative visual information.

\textbf{Patch Brightness Uncertainty (PBU).}
In the LiDAR-supported zone, patch-wide brightness changes can be mistaken for motion during direct visual updates.
Using an affine brightness model~\cite{dso}, we treat these changes as independent, zero-mean gain $\alpha$ and offset $b$ perturbations shared across the patch.
The residual covariance (Appendix~\ref*{sec:supp-patch-brightness}) is
\begin{equation}
  \mathbf C_{\mathrm{photo}}
  =\sigma_n^2\mathbf I
  +\sigma_\alpha^2\mathbf i_{\mathrm{ref}}\mathbf i_{\mathrm{ref}}^\mathsf T
  +\sigma_b^2\mathbf 1\mathbf 1^\mathsf T,
  \label{eq:patch-brightness-covariance}
\end{equation}
where $\mathbf i_{\mathrm{ref}}$ contains reference patch intensities,
$\mathbf 1$ is the all-ones vector, and $\sigma_n$, $\sigma_\alpha$, and $\sigma_b$
are the standard deviations of independent pixel noise, gain perturbations,
and offset perturbations, respectively.
We whiten residuals and Jacobians with this covariance to account for shared brightness uncertainty during visual updates.

\subsection{Gaussian Mapping}
\label{sec:erp-mapping}

\textbf{Gaussian Rendering.}
Using the projections in Sec.~\ref{sec:erp-camera}, we sort Gaussians by radial depth $R_g=\|\boldsymbol\mu_g^C\|_2$ and render panoramic color and depth through front-to-back alpha compositing.
At pixel $\mathbf u=(u,v)^\mathsf T$, let $a_g$ be the opacity-weighted projected Gaussian response and $T_g=\prod_{h<g}(1-a_h)$ the transmittance.
The accumulated opacity and rendered radial depth used by DAEP are
\begin{equation}
  A=\sum_g T_ga_g,\qquad
  \widehat D=\frac{\sum_g T_ga_g R_g}{A},\quad A>0.
  \label{eq:erp-compositing}
\end{equation}

\textbf{Initialization.}
For each mapping keyframe, Depth Any Panoramas (DAP)~\cite{dap} (Appendix~\ref*{sec:supp-depth-estimator}) predicts radial depth $D_P$.
We align it to LiDAR depth $D_L$ using the median depth ratio $s$ over their valid overlap.
The fused depth $D_F$ is
\begin{equation}
  D_F(\mathbf u)=
  \begin{cases}
    D_L(\mathbf u), & D_L(\mathbf u)>0,\\
    sD_P(\mathbf u), & D_L(\mathbf u)=0,\ D_P(\mathbf u)>0,\\
    0, & \text{otherwise}.
  \end{cases}
  \label{eq:fused-depth}
\end{equation}
Sampled LiDAR points directly initialize Gaussians.
For each patch in the LiDAR blind zone, we apply depth-continuity filtering to $D_F$ and back-project the pixel with minimum valid depth via Eq.~\eqref{eq:erp-projection} to initialize a Gaussian.

\begin{figure}[!htbp]
\centering
\includegraphics[width=0.9\linewidth]{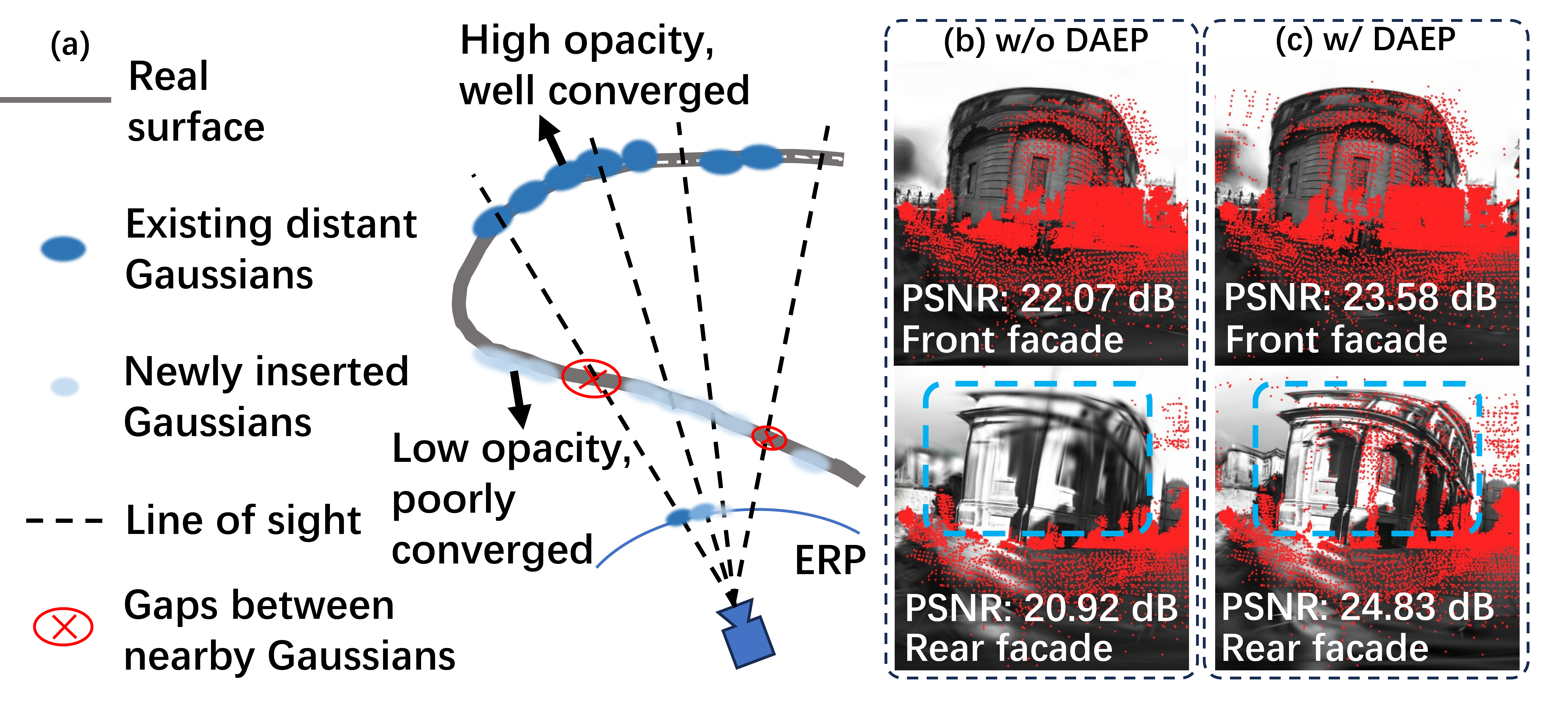}
\caption{\textbf{Depth-Aware Map Extension and Pruning (DAEP)}.
(a) Opacity-only insertion can miss new surfaces.
(b,c) Renderings without and with DAEP. Red dots mark the initial centers of Gaussians added over five neighboring keyframes.
Blue dashed boxes highlight the rear facade.}
\label{fig:depth_aware_extension}
\end{figure}

\textbf{Depth-Aware Map Extension and Pruning (DAEP).}
Methods such as PanoGS-SLAM~\cite{panogs_slam} increase Gaussian insertion density in regions of low accumulated opacity.
However, high accumulated opacity from existing Gaussians can inhibit insertion on newly observed nearby surfaces (Fig.~\ref{fig:depth_aware_extension}).
Low opacity and incomplete coverage of new Gaussians can also leave distant surfaces visible.
To address these issues, DAEP combines depth-aware insertion ~\cite{splatam}~\cite{gsslam}, with depth-conflict pruning.
A valid initialization candidate at pixel $\mathbf u$ with depth $d$ is inserted if
\begin{equation}
  A(\mathbf u)<\tau_\alpha
  \quad\text{or}\quad
  \widehat D(\mathbf u)-d>\max(\delta_0,\beta d).
  \label{eq:depth-aware-extension}
\end{equation}
The first condition fills uncovered regions, while the second admits nearer observations when the rendered surface lies behind them, even at high accumulated opacity.
Here, $\tau_\alpha$ is the opacity threshold, while $\delta_0$ and $\beta$ control the absolute and relative depth tolerances, respectively.
For depth-conflict insertions, we use higher initial opacity and scale and adaptively weight the Adam updates according to depth discrepancies:
\begin{equation}
  \Delta\boldsymbol{\theta}_g=w_g\Delta\boldsymbol{\theta}_g^{\mathrm{Adam}},\qquad w_g\ge1,
  \label{eq:daep-weighted-updates}
\end{equation}
where $\Delta\boldsymbol{\theta}_g^{\mathrm{Adam}}$ is Gaussian $g$'s standard Adam update and $w_g$ is a depth-guided adaptive weight.

To remove floating artifacts, each mapping round checks Gaussian depth consistency against the current keyframe.
Gaussians whose opacity exceeds a threshold and whose centers lie in front of the observed surface by more than a depth tolerance are pruned when local depth continuity confirms the conflict.
The check prioritizes LiDAR depth and uses DAP estimates with a larger conflict tolerance in the LiDAR blind zone.
We then optimize Gaussian parameters while keeping frontend poses fixed:
\begin{equation}
  \mathcal L_{\mathrm{map}}
  =(1-\lambda_s)\mathcal L_{\mathrm{RGB}}
  +\lambda_s(1-\operatorname{SSIM})
  +\lambda_d\mathcal L_{\mathrm{depth}}.
  \label{eq:mapping-objective}
\end{equation}
Here, $\mathcal L_{\mathrm{RGB}}$ is the mean absolute RGB error, averaged across the selected optimization views.
$\mathcal L_{\mathrm{depth}}$ is the mean absolute radial-depth error at valid LiDAR pixels, and $\lambda_s$ and $\lambda_d$ are weights.
Geometry in the LiDAR blind zone is refined through multi-view photometric supervision.

\section{Experiments}
\label{sec:experiments}

\subsection{The WHOLE Dataset}
\label{sec:whole-dataset}

We build the \textbf{WHOLE} dataset with UAV motion under challenging geometric and illumination conditions, covering \underline{W}eak-geometry, \underline{H}igh-dynamic, \underline{O}mnidirectional, \underline{L}ow-light, and \underline{E}asy scenarios. It contains 20 sequences of panoramic images, LiDAR scans, and IMU measurements collected in indoor and outdoor campus environments.
The scenes include \textit{rooms}, \textit{long corridors}, \textit{garages}, \textit{glass facades}, and \textit{open spaces} (Fig.~\ref{fig:self-made-dataset}).
Recordings span daytime and nighttime conditions, low-light garages, and rapid motion with image blur.
Some sequences combine these visual challenges with weak geometry, enabling evaluation of localization and reconstruction under simultaneous visual and geometric degradation. Data are collected using a customized FanciSwarm Pro~\cite{fanciswarmpro} UAV platform
(Fig.~\ref{fig:system-overview}).
The platform carries a MID360S LiDAR with its built-in IMU and an Insta360 X5 dual-fisheye panoramic camera.
We use Kalibr~\cite{furgale2013unified} with an AprilGrid to calibrate the intrinsics of both fisheye cameras and the camera--IMU spatial and temporal parameters.
LiDAR--camera extrinsics are calibrated with FAST-Calib2~\cite{fastcalib2}.
AprilGrid observations provide reference poses at the sequence endpoints.

\begin{figure}[!htbp]
\centering
\includegraphics[width=0.88\linewidth]{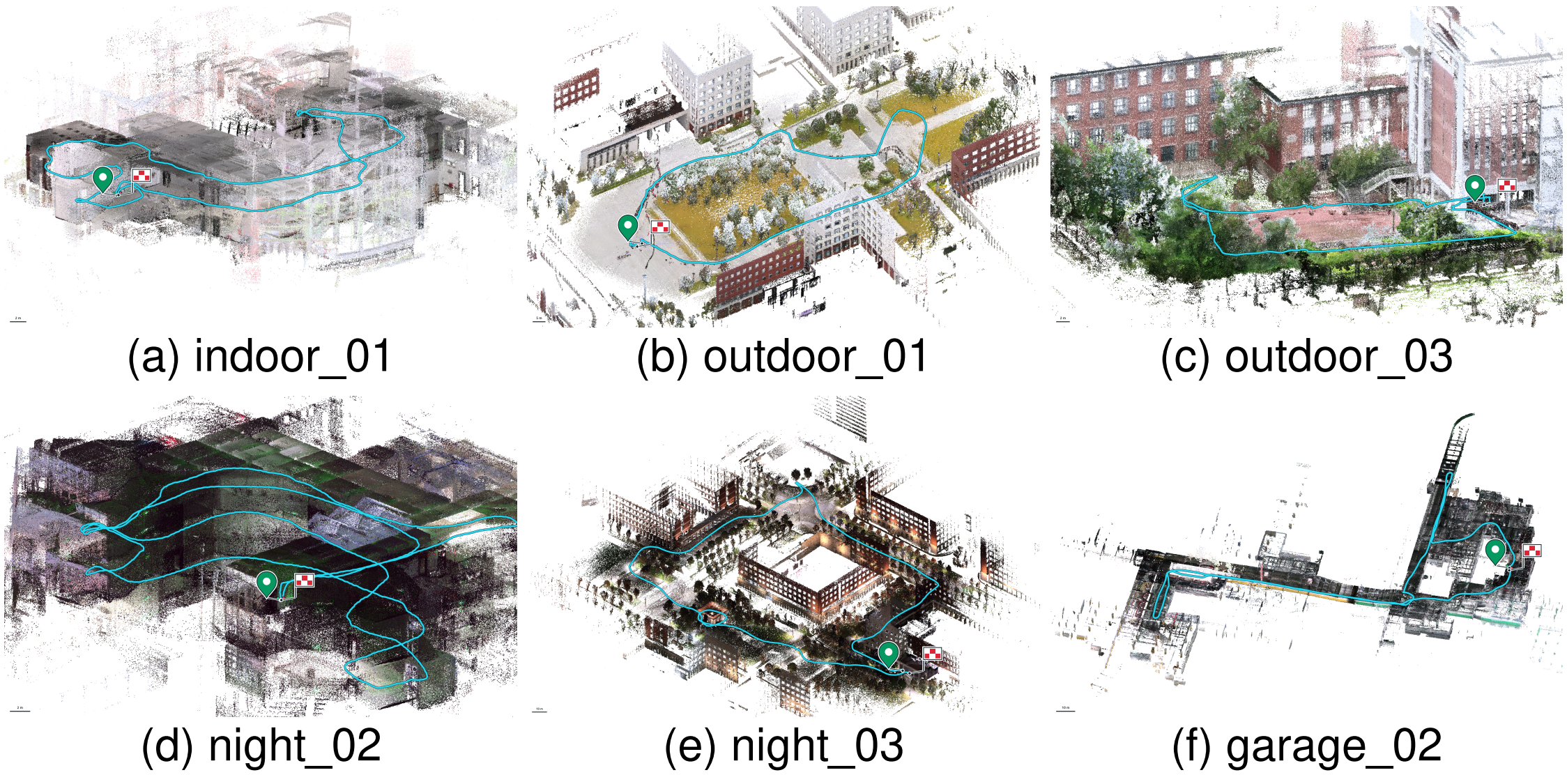}
\caption{\textbf{Visualization of scenes from WHOLE}: (a) indoor, (b,c) outdoor, (d,e) nighttime, and (f) garage.}
\label{fig:self-made-dataset}
\end{figure}

\subsection{Experimental Setup}
\label{sec:experimental-setup}

\textbf{Datasets.}
In this work, we conduct experiments on five datasets. Table~\ref{tab:datasets} summarizes the statistics and comparison between existing four datasets and WHOLE.
From the Hilti-Oxford 2022 Dataset (\textbf{Hilti-Oxford})~\cite{hiltioxford}, we select sequences covering a construction site (Exp04), a basement (Exp14), a corridor and gallery (Exp18), and building exteriors (Exp21).
\textbf{M3DGR}~\cite{m3dgr} provides sequences with varying illumination (Varying-illu05), long corridors (Corridor01/02), and outdoor grassy areas (Grass01/02).
On M3DGR, GOLD-SLAM and Gaussian-LIC2 use MID-360 as their sole LiDAR input.
\textbf{M2DGR}~\cite{m2dgr} covers indoor rooms and halls (room\_01 and hall\_02), an outdoor-to-indoor transition (door\_02), and an outdoor gate area (gate\_03).
From the Oxford Spires Dataset (\textbf{Oxford Spires})~\cite{oxfordspires}, we select Keble03, Observatory01, ChristChurch03, and Blenheim01, covering architectural interiors and outdoor spaces.

\begin{table}[!htbp]
\caption{\textbf{Comparison of public SLAM datasets and our proposed WHOLE dataset}. W: weak geometry; H: high dynamics; O: omnidirectional coverage; L: low light; E: easy scenarios.}
\label{tab:datasets}
\centering
\begingroup
\fontsize{9}{11}\selectfont
\setlength{\tabcolsep}{2pt}
\renewcommand{\arraystretch}{1.0}
\newlength{\datasettablewidth}
\setlength{\datasettablewidth}{\dimexpr\textwidth-48.5pt\relax}
\newcommand{\challengeyes}{\makebox[1em][l]{\rlap{$\square$}\hspace{0.12em}\smash{\raisebox{0.2ex}{\ding{51}}}}}
\newcommand{\challengeno}{\textemdash}
\begin{tabular}{@{}>{\raggedright\arraybackslash}p{\dimexpr.09\datasettablewidth+19.5pt\relax}
>{\centering\arraybackslash}p{\dimexpr.045\datasettablewidth+1.5pt\relax}
>{\centering\arraybackslash}p{\dimexpr.045\datasettablewidth+1.5pt\relax}
>{\raggedright\arraybackslash}p{.09\datasettablewidth}
>{\raggedright\arraybackslash}p{\dimexpr.20\datasettablewidth-4pt\relax}
>{\centering\arraybackslash}p{.145\datasettablewidth}
>{\centering\arraybackslash}p{\dimexpr.045\datasettablewidth-4pt\relax}
>{\raggedright\arraybackslash}p{.05\datasettablewidth}
>{\raggedright\arraybackslash}p{.15\datasettablewidth}
*{4}{>{\centering\arraybackslash}p{\dimexpr.028\datasettablewidth-2.7pt\relax}@{\hspace{1.5pt}}}
>{\centering\arraybackslash}p{\dimexpr.028\datasettablewidth-2.7pt\relax}@{}}
\toprule
\multirow{2}{*}{\textbf{Dataset}} & \multirow{2}{*}{\textbf{Year}} & \multirow{2}{*}{\textbf{\#Seq.}} & \multirow{2}{*}{\textbf{Platform}} & \multicolumn{3}{c}{\textbf{Camera}} & \multirow{2}{*}{\shortstack{\textbf{IMU}\\(Hz)}} & \multirow{2}{*}{\textbf{LiDAR}} & \multicolumn{5}{c}{\textbf{Data Diversity}} \\
\cmidrule(lr){5-7}\cmidrule(lr){10-14}
& & & & Type & Resolution & FPS & & & W & H & O & L & E \\
\midrule
M2DGR\cite{m2dgr} & 2021 & 36 & Ground \newline robot & $7\times$ fisheye (RGB)\newline Perspective RGB-D & $1280\!\times\!1024$\newline $640\!\times\!480$ & 15\newline 15 & 150 & VLP-32C, 10 Hz & \challengeyes & \challengeyes & \challengeyes & \challengeyes & \challengeyes \\
\midrule
Hilti-Oxford~\cite{hiltioxford} & 2022 & 16 & Handheld & $5\times$ fisheye (gray) & $720\!\times\!540$ & 40 & 400 & PandarXT-32, 10 Hz & \challengeyes & \challengeno & \challengeyes & \challengeyes & \challengeyes \\
\midrule
Oxford Spires~\cite{oxfordspires} & 2024 & 24 & Backpack & $3\times$ fisheye (RGB) & $1440\!\times\!1080$ & 20 & 400 & Hesai QT64, 10 Hz & \challengeyes & \challengeno & \challengeyes & \challengeyes & \challengeyes \\
\midrule
M3DGR\cite{m3dgr} & 2025 & 31 & Ground \newline robot & Panoramic X4 (RGB)\newline Perspective RGB-D & $2880\!\times\!1440$ (ERP)\newline $640\!\times\!480$ & 15\newline 30 & 200 & Avia, 10 Hz; MID-360, 10 Hz & \challengeyes & \challengeno & \challengeyes & \challengeyes & \challengeyes \\
\midrule
\rowcolor{gray!15} \textbf{WHOLE (Ours)} & 2026 & 20 & Handheld; UAV & Panoramic X5 (RGB) & $2\times(1920\!\times\!1920)$\newline (fisheye) & 30 & 200 & MID360S, 10 Hz & \challengeyes & \challengeyes & \challengeyes & \challengeyes & \challengeyes \\
\bottomrule
\end{tabular}
\endgroup
\end{table}

\textbf{Baselines.}
We evaluate FAST-LIO2~\cite{fastlio2}, FAST-LIVO2~\cite{fastlivo2}, and Ground-Fusion++~\cite{m3dgr} for localization, and Gaussian-LIC2~\cite{gaussianlic2}, ODGS-SLAM~\cite{odgsslam}, and PanoGS-SLAM~\cite{panogs_slam} for both localization and reconstruction.
Visual inputs are chosen to match each method's camera model and supported view configuration (Table~\ref{tab:baseline-inputs}).
Ground-Fusion++ (LIC) uses LiDAR, IMU, and front-view images, with GNSS and wheel odometry disabled.
For consistent depth inputs, ODGS-SLAM and PanoGS-SLAM use GOLD-SLAM's fused depth $D_F$.

\begin{table}[!htbp]
\caption{\textbf{Baseline inputs and evaluation tasks.} L: localization task; M: mapping task.}
\label{tab:baseline-inputs}
\centering
\small
\setlength{\tabcolsep}{2.75pt}
\begin{tabular*}{\columnwidth}{@{\extracolsep{\fill}}llccccc@{}}
\toprule
\textbf{Method} & \textbf{Visual input} & \textbf{L} & \textbf{M} & \textbf{LiDAR} & \textbf{IMU} & \textbf{Camera} \\
\midrule
FAST-LIO2 & -- & \ding{51} & \ding{55} & \ding{51} & \ding{51} & \ding{55} \\
FAST-LIVO2 & Perspective, $97^\circ$ & \ding{51} & \ding{55} & \ding{51} & \ding{51} & \ding{51} \\
Ground-Fusion++ & Perspective, $97^\circ$ & \ding{51} & \ding{55} & \ding{51} & \ding{51} & \ding{51} \\
Gaussian-LIC2 & Perspective, $97^\circ$ & \ding{51} & \ding{51} & \ding{51} & \ding{51} & \ding{51} \\ \midrule
ODGS-SLAM & ERP, $360^\circ$ & \ding{51} & \ding{51} & \ding{55} & \ding{55} & \ding{51} \\
PanoGS-SLAM & ERP, $360^\circ$ & \ding{51} & \ding{51} & \ding{55} & \ding{55} & \ding{51} \\
\textbf{GOLD-SLAM} & ERP, $360^\circ$ & \ding{51} & \ding{51} & \ding{51} & \ding{51} & \ding{51} \\
\bottomrule
\end{tabular*}
\end{table}

\textbf{Implementation Details.}
Reconstruction inputs are sampled at 10 Hz: $1024\times512$ ERP images or $512\times384$ perspective images with a $97^\circ$ horizontal FoV.
GOLD-SLAM selects a mapping keyframe every 5 frames and optimizes up to $70$ views per round (Appendices~\ref*{sec:supp-keyframe-selection} and~\ref*{sec:supp-optimization-budget}).
Gaussian colors use 3rd-order SH; loss weights are $\lambda_s=0.2$ and $\lambda_d=0.015$.
Map extension parameters are $\tau_\alpha=0.99$, $\delta_0=0.25\,\mathrm m$, and $\beta=0.1$.
We set $\gamma=0.28$ for blind-zone visual constraints and $\sigma_\alpha=0.05$, $\sigma_b=2$ for PBU.
Experiments run on an Intel Core i7-14700KF with 64 GB RAM and an NVIDIA RTX 5090 D v2 with 24 GB VRAM.

\subsection{Localization Accuracy}
\label{sec:localization-results}

We first evaluate localization accuracy using ATE RMSE and translational endpoint drift.
Across the public datasets, GOLD-SLAM achieves the lowest mean ATE on Hilti-Oxford and M3DGR, ranks second on Oxford Spires, and provides comparable accuracy on M2DGR (Table~\ref{tab:odometry}).

\begin{table}[!htbp]
\caption{\textbf{Localization accuracy} (ATE RMSE in m, $\downarrow$). Values are means (sample standard deviations).
Best and second-best results are bold (darker shading) and underlined (lighter shading), respectively.}
\label{tab:odometry}
\centering
\small
\newcommand{\atestat}[2]{\mbox{$#1$\nobreakspace{}\raisebox{-1.5pt}{\fontsize{8}{9}\selectfont\textcolor{black!60}{(#2)}}}}
\setlength{\tabcolsep}{2.75pt}
\renewcommand{\arraystretch}{1.05}
\begin{tabularx}{\columnwidth}{@{}l*{4}{Y}@{}}
\toprule
\textbf{Method} & \shortstack{Hilti-\\Oxford} & M3DGR & M2DGR & \shortstack{Oxford\\Spires} \\
\midrule
FAST-LIO2
& \atestat{1.865}{3.397}
& \atestat{0.351}{0.227}
& \atestat{0.358}{0.130}
& \cellcolor{rankBest}\atestat{\mathbf{0.075}}{0.047} \\
\addlinespace[2pt]
FAST-LIVO2
& \cellcolor{rankSecond}\atestat{\underline{0.064}}{0.033}
& \cellcolor{rankSecond}\atestat{\underline{0.346}}{0.238}
& \atestat{0.360}{0.119}
& \atestat{0.100}{0.057} \\
\addlinespace[2pt]
Gaussian-LIC2
& \atestat{0.196}{0.132}
& \atestat{0.379}{0.183}
& \cellcolor{rankSecond}\atestat{\underline{0.355}}{0.135}
& \atestat{0.563}{0.221} \\
\addlinespace[2pt]
Ground-Fusion++ (LIC)
& \atestat{7.693}{15.070}
& \atestat{0.380}{0.199}
& \cellcolor{rankBest}\atestat{\mathbf{0.349}}{0.137}
& \atestat{0.337}{0.453} \\
\addlinespace[2pt]
ODGS-SLAM
& \atestat{9.165}{4.223}
& \atestat{3.231}{1.881}
& \atestat{2.316}{2.189}
& \atestat{5.360}{0.489} \\
PanoGS-SLAM
& \atestat{9.167}{4.356}
& \atestat{29.316}{29.053}
& \atestat{12.627}{17.066}
& \atestat{32.845}{16.454} \\
\addlinespace[2pt]
\textbf{GOLD-SLAM (Ours)}
& \cellcolor{rankBest}\atestat{\mathbf{0.059}}{0.028}
& \cellcolor{rankBest}\atestat{\mathbf{0.344}}{0.241}
& \atestat{0.360}{0.119}
& \cellcolor{rankSecond}\atestat{\underline{0.077}}{0.057} \\
\bottomrule
\end{tabularx}
\end{table}

Long corridors challenge both LiDAR registration and visual tracking through repetitive geometry and appearance.
In Table~\ref{tab:sparse-trajectory}, FAST-LIO2 accumulates large drift on both corridors, while FAST-LIVO2 fails on Corridor01.
On Corridor02, GOLD-SLAM reduces endpoint drift from $1.319\,\mathrm m$ with FAST-LIVO2 to $0.274\,\mathrm m$.
This improved drift control may partly stem from additional visual constraints provided by panoramic observations beyond the forward view, including those in the LiDAR blind zone.

WHOLE extends this evaluation to weak geometry and rapid motion.
GOLD-SLAM achieves the lowest mean endpoint drift in groups W and H, with reductions of 34\% and 43\% relative to FAST-LIVO2 (Table~\ref{tab:sparse-trajectory}).
Broad visual coverage can preserve overlap as the viewing direction changes, while PBU accounts for shared brightness variations in visual updates.
The Exp18 ablation further supports the contributions of blind-zone updates and PBU: removing either increases ATE by 61\% and 58\%, respectively (Table~\ref{tab:System ablation}).

\begin{table}[!htbp]
\caption{\textbf{Translational endpoint drift} (m, $\downarrow$). M3DGR provides only endpoint ground truth for Corridor01/02; the other three selected sequences are evaluated in Table~\ref{tab:odometry}. For WHOLE, values are groupwise means (sample standard deviations). A WHOLE sequence may belong to multiple groups. \textcolor{red}{$*$} indicates failure.}
\label{tab:sparse-trajectory}
\centering
\small
\newcommand{\driftstd}[1]{\nobreakspace{}\raisebox{-1.5pt}{\fontsize{8}{9}\selectfont\textcolor{black!60}{(#1)}}}
\setlength{\tabcolsep}{1.5pt}
\begin{tabularx}{\textwidth}{@{}l
*{2}{>{\hsize=.76\hsize\linewidth=\hsize}Y}
>{\hsize=1.22\hsize\linewidth=\hsize}Y
*{2}{>{\hsize=1.02\hsize\linewidth=\hsize}Y}
>{\hsize=1.22\hsize\linewidth=\hsize}Y@{}}
\toprule
\multirow{2}{*}{\textbf{Method}}
& \multicolumn{2}{c}{\textbf{M3DGR}}
& \multicolumn{4}{c}{\textbf{WHOLE (Ours)}} \\
\cmidrule(lr){2-3}\cmidrule(lr){4-7}
& \shortstack{Corridor01} & \shortstack{Corridor02}
& \textbf{W}eak-geometry & \textbf{H}igh-dynamic & \textbf{L}ow-light & \textbf{E}asy \\
\midrule
FAST-LIO2
& $18.317$
& $17.491$
& $3.868$\driftstd{5.349}
& $2.025$\driftstd{3.998}
& $\rankbest{1.039}$\driftstd{2.541}
& $2.746$\driftstd{6.069} \\
\addlinespace[4pt]
FAST-LIVO2
& $34.548^{\textcolor{red}{*}}$
& $\ranksecond{1.319}$
& $\ranksecond{2.719}$\driftstd{3.690}
& $\ranksecond{1.806}$\driftstd{3.516}
& $1.418$\driftstd{2.456}
& $0.066$\driftstd{0.116} \\
\addlinespace[4pt]
Gaussian-LIC2
& $14.272$
& $19.512$
& $3.069$\driftstd{3.927}
& $1.988$\driftstd{3.738}
& $1.712$\driftstd{2.682}
& $\rankbest{0.020}$\driftstd{0.007} \\
\addlinespace[4pt]
Ground-Fusion++ (LIC)
& $\ranksecond{2.700}$
& $5.065$
& $135.126^{\textcolor{red}{*}}$\driftstd{367.835}
& $18.405$\driftstd{60.137}
& $1.753$\driftstd{3.303}
& $245.128^{\textcolor{red}{*}}$\driftstd{548.105} \\
\addlinespace[4pt]
ODGS-SLAM
& $11.309$
& $9.252$
& $35.309^{\textcolor{red}{*}}$\driftstd{31.618}
& $22.215^{\textcolor{red}{*}}$\driftstd{21.939}
& $27.858^{\textcolor{red}{*}}$\driftstd{35.408}
& $8.044$\driftstd{6.632} \\
PanoGS-SLAM
& $49.698^{\textcolor{red}{*}}$
& $10.223$
& $16.895$\driftstd{16.824}
& $12.578$\driftstd{16.043}
& $16.297$\driftstd{19.311}
& $5.709$\driftstd{2.762} \\
\addlinespace[4pt]
\textbf{GOLD-SLAM (Ours)}
& $\rankbest{1.161}$
& $\rankbest{0.274}$
& $\rankbest{1.808}$\driftstd{2.276}
& $\rankbest{1.021}$\driftstd{2.060}
& $\ranksecond{1.302}$\driftstd{2.041}
& $\ranksecond{0.065}$\driftstd{0.116} \\
\bottomrule
\end{tabularx}
\end{table}

\subsection{Rendering and Geometric Reconstruction}
\label{sec:mapping-results}
We evaluate in-sequence novel-view rendering and depth consistency on shared front views, with per-sequence results provided in Appendix~\ref*{sec:supp-sequence-evaluation}.
In Table~\ref{tab:front-rendering}, GOLD-SLAM achieves the lowest mean Depth-L1 across all four public datasets and all four WHOLE groups, reducing the error by 20--33\% relative to Gaussian-LIC2 across WHOLE.
GOLD-SLAM also leads most rendering comparisons, generally outperforming Gaussian-LIC2.
GOLD-SLAM improves mean PSNR by 9.10 and 5.56\,dB and reduces mean Depth-L1 by 71.9\% and 83.9\% relative to ODGS-SLAM and PanoGS-SLAM, respectively.
This geometric advantage persists even though both panoramic baselines receive GOLD-SLAM's fused depth $D_F$.
Beyond depth completion, GOLD-SLAM uses measured LiDAR depth to supervise map optimization.
The DAP-only ablation in Table~\ref{tab:System ablation} supports the role of measured depth: replacing LiDAR measurements with predictions increases reconstruction error under the same frontend poses.

\Needspace{5\baselineskip}
Higher image fidelity need not imply more accurate surface geometry.
For example, on Grass02, PanoGS-SLAM achieves higher PSNR than GOLD-SLAM (22.87 vs.\ 18.92 dB), but also higher Depth-L1 (11.312 vs.\ 6.980 m; see Appendix~\ref*{sec:supp-sequence-evaluation}).

\begin{table}[!htbp]
\caption{\textbf{Results of in-sequence novel-view rendering and depth consistency} on shared $512\times384$ front views ($97^\circ$ horizontal FoV).
Values are means (sample standard deviations) over sequences; Depth-L1 is radial-depth MAE against LiDAR.}
\label{tab:front-rendering}
\centering
\small
\newcommand{\renderstd}[1]{\nobreakspace{}\raisebox{-1.5pt}{\fontsize{8}{9}\selectfont\textcolor{black!60}{(#1)}}}
\renewcommand{\arraystretch}{1.03}
\setlength{\tabcolsep}{3pt}
\begin{tabularx}{\textwidth}{@{}ll*{4}{Y}@{}}
\toprule
\multirow{2}{*}{Dataset} & \multirow{2}{*}{Method} & \multicolumn{3}{c}{\textbf{Rendering Quality}} & \multicolumn{1}{c}{\textbf{Geometric Accuracy}} \\
\cmidrule(lr){3-5}\cmidrule(lr){6-6}
& & PSNR $\uparrow$ (dB) & SSIM $\uparrow$ & LPIPS $\downarrow$ & Depth-L1 $\downarrow$ (m) \\
\midrule
\multirow{4}{*}{\textbf{Hilti-Oxford 2022}}
 & Gaussian-LIC2 & $\ranksecond{26.31}$\renderstd{1.90} & $\ranksecond{0.895}$\renderstd{0.028} & $\ranksecond{0.236}$\renderstd{0.068} & $\ranksecond{1.052}$\renderstd{0.692} \\
 & ODGS-SLAM & $13.88$\renderstd{1.73} & $0.493$\renderstd{0.114} & $0.610$\renderstd{0.031} & $4.653$\renderstd{3.503} \\
 & PanoGS-SLAM & $19.12$\renderstd{1.35} & $0.776$\renderstd{0.050} & $0.528$\renderstd{0.046} & $5.512$\renderstd{4.149} \\
 & \textbf{GOLD-SLAM (Ours)} & $\rankbest{27.13}$\renderstd{1.61} & $\rankbest{0.916}$\renderstd{0.024} & $\rankbest{0.186}$\renderstd{0.070} & $\rankbest{0.869}$\renderstd{0.718} \\
\midrule
\multirow{4}{*}{\textbf{M3DGR}}
 & Gaussian-LIC2 & $\ranksecond{22.76}$\renderstd{3.69} & $0.726$\renderstd{0.172} & $\ranksecond{0.332}$\renderstd{0.086} & $\ranksecond{3.687}$\renderstd{3.045} \\
 & ODGS-SLAM & $19.96$\renderstd{3.03} & $0.621$\renderstd{0.118} & $0.364$\renderstd{0.043} & $6.132$\renderstd{5.465} \\
 & PanoGS-SLAM & $22.38$\renderstd{1.89} & $\rankbest{0.732}$\renderstd{0.131} & $0.438$\renderstd{0.046} & $7.602$\renderstd{3.824} \\
 & \textbf{GOLD-SLAM (Ours)} & $\rankbest{23.39}$\renderstd{4.07} & $\ranksecond{0.730}$\renderstd{0.193} & $\rankbest{0.310}$\renderstd{0.112} & $\rankbest{3.247}$\renderstd{2.982} \\
\midrule
\multirow{4}{*}{\textbf{M2DGR}}
 & Gaussian-LIC2 & $\rankbest{18.94}$\renderstd{1.07} & $\rankbest{0.683}$\renderstd{0.034} & $\rankbest{0.342}$\renderstd{0.043} & $\ranksecond{2.058}$\renderstd{1.941} \\
 & ODGS-SLAM & $13.21$\renderstd{1.32} & $0.365$\renderstd{0.061} & $0.570$\renderstd{0.056} & $3.735$\renderstd{2.157} \\
 & PanoGS-SLAM & $15.95$\renderstd{0.89} & $0.570$\renderstd{0.057} & $0.509$\renderstd{0.031} & $11.659$\renderstd{9.770} \\
 & \textbf{GOLD-SLAM (Ours)} & $\ranksecond{18.72}$\renderstd{2.38} & $\ranksecond{0.670}$\renderstd{0.119} & $\ranksecond{0.347}$\renderstd{0.088} & $\rankbest{1.653}$\renderstd{1.649} \\
\midrule
\multirow{4}{*}{\textbf{Oxford Spires}}
 & Gaussian-LIC2 & $\ranksecond{24.94}$\renderstd{1.27} & $\ranksecond{0.841}$\renderstd{0.038} & $\ranksecond{0.256}$\renderstd{0.047} & $\ranksecond{1.227}$\renderstd{0.352} \\
 & ODGS-SLAM & $15.58$\renderstd{5.77} & $0.589$\renderstd{0.179} & $0.584$\renderstd{0.071} & $2.534$\renderstd{0.368} \\
 & PanoGS-SLAM & $20.26$\renderstd{1.10} & $0.732$\renderstd{0.051} & $0.624$\renderstd{0.065} & $8.356$\renderstd{1.546} \\
 & \textbf{GOLD-SLAM (Ours)} & $\rankbest{26.15}$\renderstd{1.58} & $\rankbest{0.880}$\renderstd{0.029} & $\rankbest{0.164}$\renderstd{0.035} & $\rankbest{0.514}$\renderstd{0.158} \\
\midrule
\multirow{4}{*}{\textbf{WHOLE-W}}
 & Gaussian-LIC2 & $\ranksecond{24.70}$\renderstd{1.54} & $\ranksecond{0.855}$\renderstd{0.042} & $\ranksecond{0.192}$\renderstd{0.057} & $\ranksecond{1.652}$\renderstd{0.910} \\
 & ODGS-SLAM & $15.73$\renderstd{1.74} & $0.550$\renderstd{0.061} & $0.526$\renderstd{0.084} & $5.870$\renderstd{4.182} \\
 & PanoGS-SLAM & $18.54$\renderstd{1.47} & $0.714$\renderstd{0.060} & $0.524$\renderstd{0.093} & $9.169$\renderstd{4.389} \\
 & \textbf{GOLD-SLAM (Ours)} & $\rankbest{25.14}$\renderstd{0.91} & $\rankbest{0.862}$\renderstd{0.040} & $\rankbest{0.158}$\renderstd{0.034} & $\rankbest{1.312}$\renderstd{0.842} \\
\midrule
\multirow{4}{*}{\textbf{WHOLE-H}}
 & Gaussian-LIC2 & $\ranksecond{25.66}$\renderstd{3.68} & $\ranksecond{0.853}$\renderstd{0.058} & $\ranksecond{0.194}$\renderstd{0.066} & $\ranksecond{1.625}$\renderstd{0.892} \\
 & ODGS-SLAM & $16.11$\renderstd{2.74} & $0.521$\renderstd{0.072} & $0.523$\renderstd{0.093} & $5.945$\renderstd{3.524} \\
 & PanoGS-SLAM & $20.08$\renderstd{3.53} & $0.734$\renderstd{0.081} & $0.478$\renderstd{0.115} & $9.647$\renderstd{4.128} \\
 & \textbf{GOLD-SLAM (Ours)} & $\rankbest{26.48}$\renderstd{3.24} & $\rankbest{0.867}$\renderstd{0.051} & $\rankbest{0.156}$\renderstd{0.040} & $\rankbest{1.294}$\renderstd{0.802} \\
\midrule
\multirow{4}{*}{\textbf{WHOLE-L}}
 & Gaussian-LIC2 & $\rankbest{27.58}$\renderstd{3.77} & $\ranksecond{0.896}$\renderstd{0.034} & $\ranksecond{0.138}$\renderstd{0.034} & $\ranksecond{1.097}$\renderstd{0.576} \\
 & ODGS-SLAM & $17.31$\renderstd{2.53} & $0.531$\renderstd{0.060} & $0.514$\renderstd{0.084} & $4.068$\renderstd{2.839} \\
 & PanoGS-SLAM & $20.68$\renderstd{4.19} & $0.764$\renderstd{0.087} & $0.457$\renderstd{0.131} & $7.059$\renderstd{2.097} \\
 & \textbf{GOLD-SLAM (Ours)} & $\ranksecond{27.55}$\renderstd{3.85} & $\rankbest{0.900}$\renderstd{0.032} & $\rankbest{0.128}$\renderstd{0.027} & $\rankbest{0.875}$\renderstd{0.593} \\
\midrule
\multirow{4}{*}{\textbf{WHOLE-E}}
 & Gaussian-LIC2 & $\ranksecond{26.19}$\renderstd{0.47} & $\ranksecond{0.875}$\renderstd{0.033} & $\ranksecond{0.173}$\renderstd{0.021} & $\ranksecond{1.067}$\renderstd{0.462} \\
 & ODGS-SLAM & $17.66$\renderstd{1.28} & $0.646$\renderstd{0.059} & $0.443$\renderstd{0.081} & $2.685$\renderstd{1.395} \\
 & PanoGS-SLAM & $21.24$\renderstd{1.57} & $0.761$\renderstd{0.063} & $0.428$\renderstd{0.060} & $7.937$\renderstd{4.448} \\
 & \textbf{GOLD-SLAM (Ours)} & $\rankbest{27.51}$\renderstd{0.54} & $\rankbest{0.894}$\renderstd{0.033} & $\rankbest{0.132}$\renderstd{0.021} & $\rankbest{0.718}$\renderstd{0.400} \\
\bottomrule
\end{tabularx}
\end{table}

\textbf{Qualitative Evaluation.}
In Fig.~\ref{fig:rendering-comparison}, Gaussian-LIC2 preserves detail in the central view, but its reconstructions become less clear toward the periphery.
ODGS-SLAM and PanoGS-SLAM exhibit blurred textures and poorly defined structures.
In contrast, GOLD-SLAM maintains sharper facade boundaries and clearer pavement textures across the wider FoV.
The panoramas in Fig.~\ref{fig:panoramic-rendering-examples} show these improvements beyond the forward view.

\begin{figure}[!htbp]
\centering
\includegraphics[width=\textwidth]{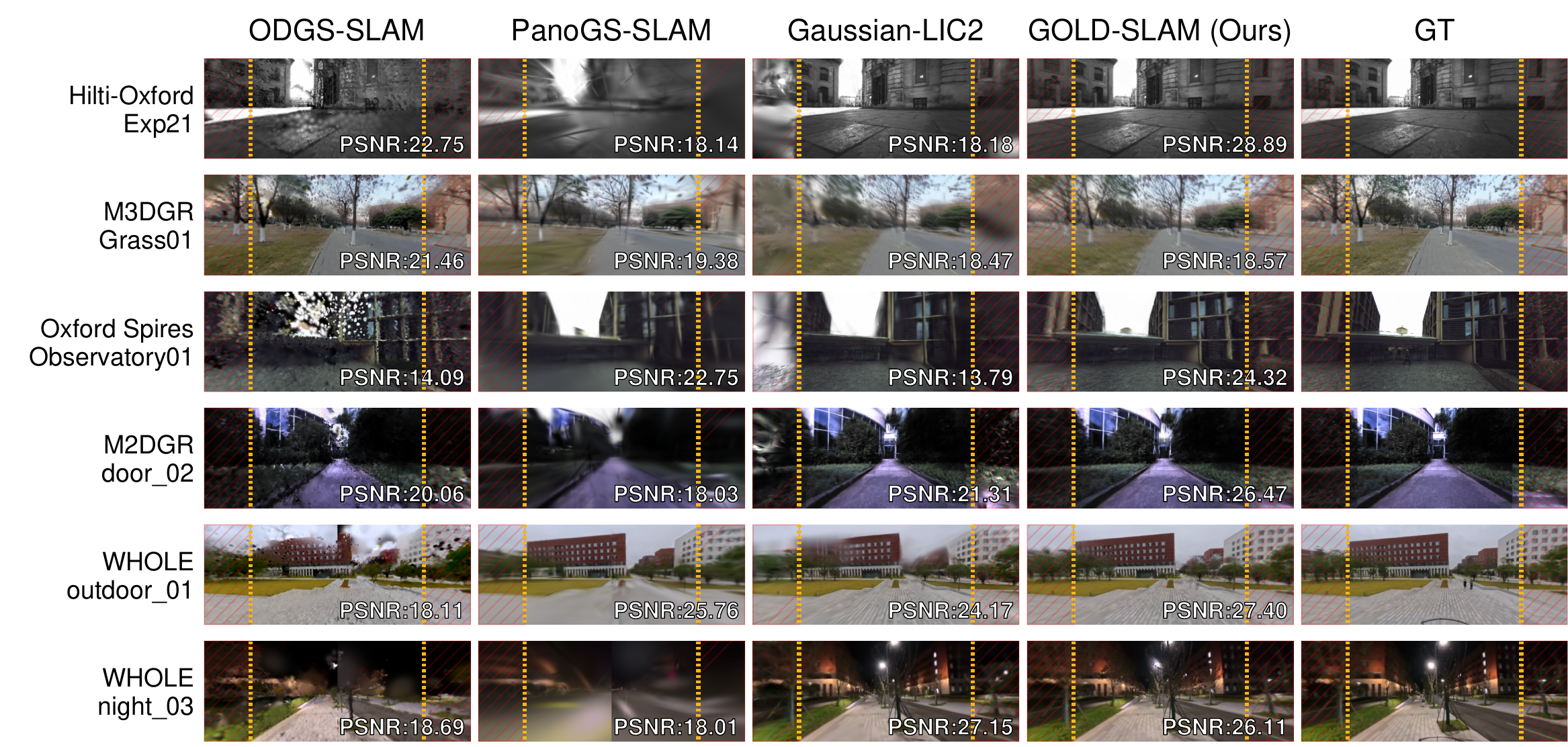}
\caption{\textbf{Qualitative comparison of scene reconstruction.}
  All panels show a $120^\circ$ horizontal FoV.
  Panoramic renderings are projected to the same perspective view.
  Yellow dashed lines mark the central $97^\circ$ horizontal FoV.}
\label{fig:rendering-comparison}
\end{figure}

\begin{figure}[!htbp]
\centering
\includegraphics[width=0.92\linewidth]{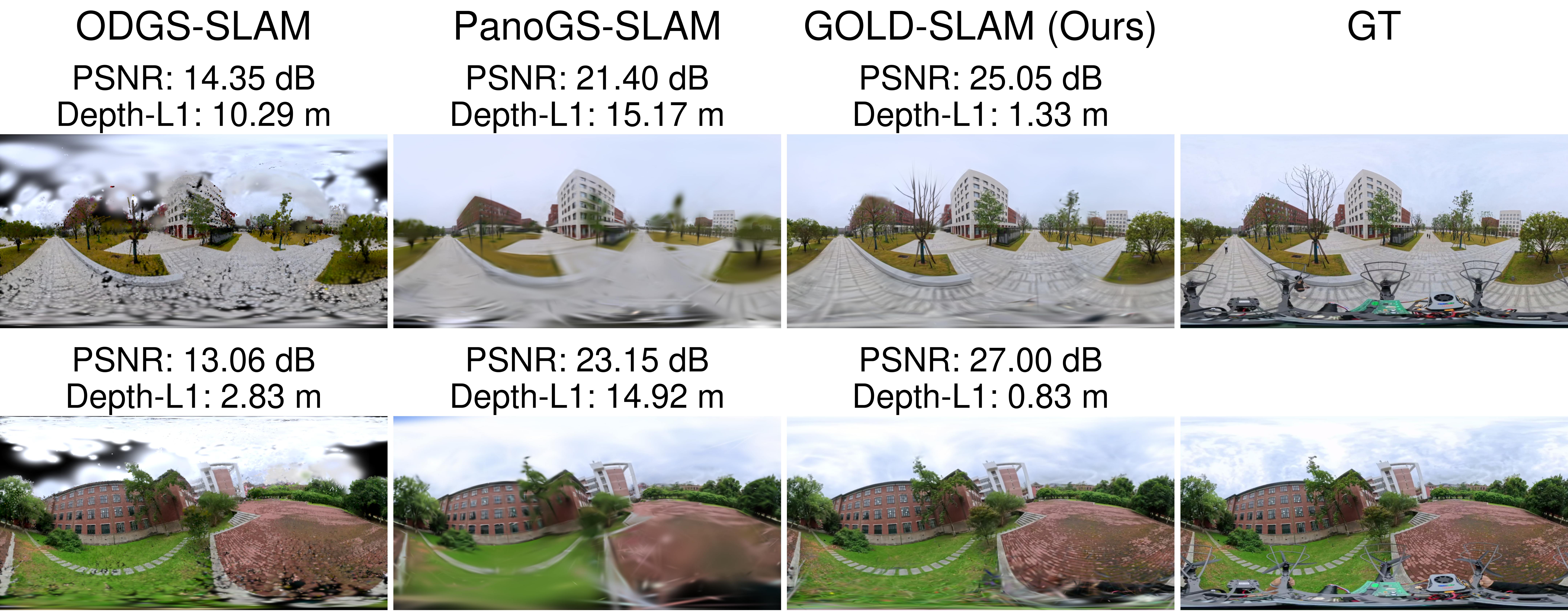}
\caption{\textbf{Panoramic reconstruction comparison} on WHOLE outdoor\_01 (top) and WHOLE outdoor\_02 (bottom).}
\label{fig:panoramic-rendering-examples}
\end{figure}

\begingroup
\clubpenalty=10000
\textbf{Computational Efficiency.}
In Table~\ref{tab:efficiency}, GOLD-SLAM takes 95.4--201.6 s, less than ODGS-SLAM and PanoGS-SLAM on both sequences.
Gaussian-LIC2 is faster, but produces smaller maps on outdoor\_01 and Keble03.
On Keble03, GOLD-SLAM uses 1.407 million Gaussians versus 0.407 million for Gaussian-LIC2, illustrating the additional mapping cost of broader scene coverage.
In addition, GOLD-SLAM supports online deployment on a Jetson Orin NX (16\,GB) (Appendix~\ref*{sec:supp-embedded-deployment}), retaining comparable localization accuracy on the completed WHOLE sequences, with a 5.80\,dB reduction in overall mean PSNR relative to the main experimental configuration.
\par
\endgroup

\begin{table}[!htbp]
\caption{\textbf{Computational cost comprison} on the first 600 input frames of each sequence.
D, FE, and BE denote depth estimation, odometry, and mapping.
$N_{\rm GS}$ denotes the final Gaussian count.}
\label{tab:efficiency}
\centering
\small
\setlength{\tabcolsep}{2.75pt}
\begin{tabularx}{\columnwidth}{@{}l*{7}{Y}@{}}
\toprule
Method & \shortstack{RAM\\{[GiB]}} & \shortstack{VRAM\\{[GiB]}}
& \shortstack{$T_{\mathrm{D}}$\\{[s]}}
& \shortstack{$T_{\mathrm{FE}}$\\{[s]}} & \shortstack{$T_{\mathrm{BE}}$\\{[s]}}
& \shortstack{Total\\{[s]}} & \shortstack{$N_{\rm GS}$\\$[10^6]$} \\
\midrule
\rowcolor{gray!15}\multicolumn{8}{l}{\textbf{\textit{outdoor\_01} randomly selected from WHOLE}} \\
Gaussian-LIC2 & 4.60 & \textbf{1.38} & \textbf{0.7} & \textbf{18.8} & \textbf{9.3} & \textbf{20.6} & 0.049 \\
ODGS-SLAM & \textbf{4.42} & 10.98 & 58.9 & 718.0 & 765.6 & 946.6 & 0.908 \\
PanoGS-SLAM & \underline{4.47} & 4.29 & \underline{4.7} & 154.6 & 290.1 & 318.3 & 0.042 \\
\textbf{GOLD-SLAM} & 7.56 & \underline{3.52} & 11.6 & \underline{20.6} & \underline{81.9} & \underline{95.4} & 0.500 \\
\midrule
\rowcolor{gray!15}\multicolumn{8}{l}{\textbf{\textit{Keble03} randomly selected from Oxford Spires}} \\
Gaussian-LIC2 & 5.18 & \textbf{3.67} & \textbf{0.8} & \textbf{41.8} & \textbf{26.4} & \textbf{45.3} & 0.407 \\
ODGS-SLAM & \underline{4.68} & 7.22 & 58.5 & 546.8 & 641.5 & 753.1 & 0.653 \\
PanoGS-SLAM & \textbf{4.56} & \underline{4.09} & \underline{6.8} & 114.4 & 229.2 & 247.0 & 0.025 \\
\textbf{GOLD-SLAM} & 7.89 & 4.28 & 12.3 & \underline{46.7} & \underline{179.6} & \underline{201.6} & 1.407 \\
\bottomrule
\end{tabularx}
\end{table}

\subsection{Ablation Studies}
\label{sec:ablation}

\textbf{Analysis of LiDAR Blind Zone.}
On Oxford Spires Keble03, we progressively crop the vertical LiDAR FoV used by the mapping backend while keeping its horizontal FoV and the frontend unchanged.
The retained solid-angle fraction $q$ decreases from $100\%$ to $0\%$ in $10$-percentage-point steps.
At $q=0$, Gaussian initialization relies entirely on DAP, without LiDAR scale alignment or depth supervision.

As shown in Fig.~\ref{fig:blind-zone}, reducing $q$ from $100\%$ to $50\%$ decreases PSNR by only $0.33$ dB, whereas Depth-L1 increases from $0.242$ m to $0.390$ m.
Further reductions in $q$ lead to more pronounced degradation in rendering quality and geometric accuracy.
Without LiDAR support in the backend, Depth-L1 reaches $9.797$ m.

\begin{figure}[!htbp]
\centering
\includegraphics[width=0.72\linewidth]{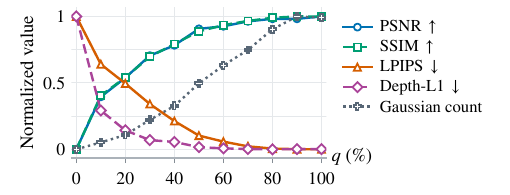}
\caption{\textbf{Analysis of LiDAR FoV on GOLD-SLAM}. Each metric is independently min--max normalized.}
\label{fig:blind-zone}
\end{figure}

\textbf{Analysis of Components.}
Table~\ref{tab:System ablation} evaluates system components on Hilti Exp18.
Disabling blind-zone visual updates or PBU ($\sigma_\alpha=\sigma_b=0$) increases ATE from 0.100 m to 0.161 m or 0.158 m, respectively, and reduces PSNR from 25.42 dB to 23.08 dB or 24.92 dB.
The three backend ablations share the frontend poses of the full model.
No DAP initializes Gaussians only from LiDAR, reducing PSNR to 24.85 dB.
DAP-only mapping uses only DAP for backend initialization and depth supervision, while its Depth-L1 rises from 0.450 m to 1.223 m.
No DAEP retains opacity-based insertion, reducing PSNR to 25.21 dB and increasing LPIPS from 0.189 to 0.214.
This supports DAEP's role in reconstructing nearer surfaces that opacity-based insertion can miss when distant Gaussians provide high opacity (Fig.~\ref{fig:depth_aware_extension}).
Full achieves the best PSNR, SSIM, LPIPS, and Depth-L1.

\begin{table}[!htbp]
\caption{
\textbf{Ablation study of GOLD-SLAM components.} Shading highlights key comparisons with the full model.}
\label{tab:System ablation}
\centering
\small
\setlength{\tabcolsep}{2.75pt}
\renewcommand{\arraystretch}{1.12}
\begin{tabularx}{\columnwidth}{@{}ll*{6}{Y}@{}}
\toprule
Metric & Unit & Full & \shortstack{w/o\\blind\\zone} & \shortstack{w/o\\PBU} & \shortstack{w/o\\DAP} & \shortstack{DAP\\only} & \shortstack{w/o\\DAEP} \\
\midrule
\multicolumn{8}{@{}l}{\textbf{(a) Accuracy}} \\
PSNR $\uparrow$ & dB & \cellcolor{gray!20}\textbf{25.42} & \cellcolor{gray!20}23.08 & \cellcolor{gray!20}24.92 & \cellcolor{gray!20}24.85 & \cellcolor{gray!20}24.61 & \cellcolor{gray!20}25.21 \\
SSIM $\uparrow$ & -- & \cellcolor{gray!20}\textbf{0.888} & \cellcolor{gray!20}0.835 & 0.876 & 0.881 & \cellcolor{gray!20}0.863 & 0.887 \\
LPIPS $\downarrow$ & -- & \cellcolor{gray!20}\textbf{0.189} & \cellcolor{gray!20}0.261 & 0.213 & 0.219 & \cellcolor{gray!20}0.244 & 0.214 \\
Depth-L1 $\downarrow$ & m & \cellcolor{gray!20}\textbf{0.450} & \cellcolor{gray!20}0.540 & \cellcolor{gray!20}0.473 & 0.451 & \cellcolor{gray!20}1.223 & \cellcolor{gray!20}0.458 \\
ATE $\downarrow$ & m & \cellcolor{gray!20}\textbf{0.100} & \cellcolor{gray!20}0.161 & \cellcolor{gray!20}0.158 & \textbf{0.100} & \textbf{0.100} & \textbf{0.100} \\
\midrule
\multicolumn{8}{@{}l}{\textbf{(b) Computational Cost}} \\
RAM & GiB & 14.76 & 11.45 & 12.88 & \textbf{10.44} & 10.80 & 11.89 \\
VRAM & GiB & 3.81 & 3.81 & 3.81 & \textbf{1.58} & 3.56 & 3.77 \\
Total & s & 184.9 & 187.2 & 182.8 & 127.9 & \textbf{117.4} & 161.8 \\
\bottomrule
\end{tabularx}
\end{table}

\FloatBarrier
\Needspace{13\baselineskip}
\section{Conclusion}
\label{sec:conclusion}

We present GOLD-SLAM, integrating omni-directional vision, LiDAR, and IMU with unified spherical tracking and Gaussian mapping.
Blind-zone visual constraints and PBU support robust pose estimation.
Measured LiDAR depth and panoramic depth predictions guide Gaussian initialization, while DAEP improves incremental surface reconstruction.
Besides, we build WHOLE, a novel multimodal dataset spanning handheld and UAV motion in challenging environments.
Experiments show competitive localization and improved geometric accuracy and rendering quality with broad scene coverage. We hope this challenging task along with its dataset can bring new insight to the community.

\noindent\textbf{Limitation and Future Work.}
Our geometric evaluation is limited to regions with LiDAR reference depth.
Future work will use simulated scenes with dense ground-truth geometry to assess blind-zone reconstruction.
We will also improve panoramic stitching and temporal alignment to reduce the effects of multi-camera parallax and synchronization errors on tracking and mapping.

\FloatBarrier
\begingroup
\raggedright
\section*{Acknowledgments}
This work was supported in part by the National Natural Science Foundation of China under Grant No. 62503166, in part by the Hunan Provincial Research and Development Project under Grant number 2026QK3018, and in part by the Project of Yuelushan Center for Industrial Innovation. This work was supported in part by Helmholtz Association of German Research Centers, in part by the Ministry of Science, Research and the Arts of Baden-Württemberg (MWK) through the Cooperative Graduate School Accessibility through AI-based Assistive Technology (KATE) under Grant BW6-03. 

\par
\endgroup
{
    \footnotesize
    \setlength{\bibsep}{2pt plus 1pt}
    \setlength{\parskip}{0pt}
    \bibliographystyle{arxiv/refs/ieeenat_fullname}
    \bibliography{arxiv/refs/main}
}
\endgroup

\clearpage
\appendix
\ReportSupplementTitle{\ReportShortTitle: \PaperSubtitle}

\section{Embedded Deployment}
\label{sec:supp-embedded-deployment}

To assess computational feasibility for UAV applications, we evaluate an online implementation of GOLD-SLAM
on a Jetson Orin NX with 16\,GB of shared memory across all 37 benchmark sequences.

\textbf{Configuration.}
We use $512\!\times\!256$ ERP/DAP inputs, FP16 DAP inference, degree-zero spherical harmonics, and 70 optimization views per update.
Sensor data are replayed at $1\times$ speed, with tracking and mapping running asynchronously.

\textbf{Evaluation and Results.}
Table~\ref{tab:supp-embedded-deployment} reports reconstruction quality, processing rates, and memory use.
Four sequences fail due to out-of-memory (OOM) errors.
The embedded configuration achieves a mean PSNR of 19.51\,dB, compared with 25.31\,dB for the configuration used in the main experiments.
Across the 17 completed WHOLE sequences, the mean endpoint drift is 0.333\,m.
Meanwhile, physical memory usage across the device peaks at 14.82\,GiB.
These results support the feasibility of embedded deployment for UAV applications, with reduced reconstruction quality and limited memory headroom under the tested configuration.

\begin{table}[!htbp]
\centering
\caption{\textbf{Reconstruction quality and computational cost on Jetson Orin NX (16\,GB).}
Entries are means with sample standard deviations.}
\label{tab:supp-embedded-deployment}
\begingroup
\fontsize{9}{11}\selectfont
\setlength{\tabcolsep}{2pt}
\newcommand{\embeddedstd}[1]{\nobreakspace{}\raisebox{-1.5pt}{\fontsize{7.5}{8.5}\selectfont\textcolor{black!60}{(#1)}}}
\renewcommand{\arraystretch}{1.0}
\begin{tabular*}{\textwidth}{@{\extracolsep{\fill}}l*{7}{c}@{}}
\toprule
\multirow{2}{*}{Dataset} & \multicolumn{4}{c}{Reconstruction} & \multicolumn{2}{c}{Processing rate} & Memory \\
\cmidrule(lr){2-5}\cmidrule(lr){6-7}\cmidrule(lr){8-8}
& PSNR [dB] & SSIM & LPIPS & Depth-L1 [m] & Front end [Hz] & Back end [keyframes/s] & RSS [GiB] \\
\midrule
Hilti-Oxford & $19.70$\embeddedstd{2.02} & $0.781$\embeddedstd{0.061} & $0.431$\embeddedstd{0.044} & $1.178$\embeddedstd{0.670} & $5.62$\embeddedstd{1.97} & $0.74$\embeddedstd{0.12} & $7.53$\embeddedstd{0.99} \\
M3DGR & $15.17$\embeddedstd{2.52} & $0.452$\embeddedstd{0.045} & $0.612$\embeddedstd{0.034} & $6.284$\embeddedstd{0.721} & $6.96$\embeddedstd{3.37} & $0.40$\embeddedstd{0.08} & $8.36$\embeddedstd{1.47} \\
M2DGR & $15.99$\embeddedstd{0.91} & $0.549$\embeddedstd{0.075} & $0.496$\embeddedstd{0.024} & $2.431$\embeddedstd{2.901} & $2.86$\embeddedstd{1.74} & $0.64$\embeddedstd{0.17} & $7.36$\embeddedstd{1.95} \\
Oxford Spires & $21.48$\embeddedstd{1.77} & $0.745$\embeddedstd{0.067} & $0.404$\embeddedstd{0.073} & $0.932$\embeddedstd{0.356} & $2.83$\embeddedstd{0.53} & $0.37$\embeddedstd{0.03} & $6.68$\embeddedstd{0.67} \\
WHOLE & $20.85$\embeddedstd{3.81} & $0.753$\embeddedstd{0.092} & $0.350$\embeddedstd{0.090} & $1.490$\embeddedstd{0.829} & $6.40$\embeddedstd{1.34} & $0.51$\embeddedstd{0.13} & $7.85$\embeddedstd{1.28} \\
\bottomrule
\end{tabular*}
\endgroup
\end{table}

\section{Spherical Geometry}
\label{sec:supp-erp-covariance}

\textbf{State and Pose Jacobian.}\phantomsection
\label{sec:supp-state-pose}
The frontend state is
\begin{equation}
  \mathbf x_t=(\mathbf R_t,\mathbf p_t,\mathbf v_t,
    \mathbf b_{g,t},\mathbf b_{a,t},\mathbf g_t).
  \label{eq:system-state}
\end{equation}
The system pose is represented by the IMU pose in the world frame, where $\mathbf R_t$ maps IMU-frame coordinates to the world frame and $\mathbf p_t$ denotes the IMU position.
The remaining state variables $\mathbf v_t$, $\mathbf b_{g,t}$, $\mathbf b_{a,t}$, and $\mathbf g_t$ denote velocity, gyroscope bias, accelerometer bias, and gravity, respectively.

For a world point $\mathbf P^W$, its coordinates in the IMU and camera frames are $\mathbf p^I=\mathbf R_t^\mathsf T(\mathbf P^W-\mathbf p_t)$ and $\mathbf p^C=\mathbf R_I^C(\mathbf p^I-\mathbf p_C^I)$, respectively, where $\mathbf p_C^I$ is the camera center expressed in the IMU frame.
For a right-multiplicative rotation perturbation and a world-frame translation increment, $\delta\boldsymbol\xi=[\delta\boldsymbol\theta^\mathsf T,\delta\mathbf p^\mathsf T]^\mathsf T$, we obtain
\begin{equation}
  \frac{\partial\pi(\mathbf p^C)}{\partial\delta\boldsymbol\xi}
  =\mathbf J_\pi\mathbf R_I^C
    \begin{bmatrix}[\mathbf p^I]^\wedge&-\mathbf R_t^\mathsf T\end{bmatrix},
  \label{eq:erp-control-jacobian}
\end{equation}
where $[\mathbf p^I]^\wedge$ is the skew-symmetric matrix of $\mathbf p^I$.

\textbf{Covariance Projection.}\phantomsection
\label{sec:supp-tangent-covariance}
Eq.~(\ref*{eq:covariance-projection}) follows from the local tangent-plane construction~\cite{odgs,odgsslam}.
For Gaussian $g$, let $R_g=\|\boldsymbol\mu_g^C\|_2$ and $\mathbf d_g=\boldsymbol\mu_g^C/R_g$ denote its radial distance and unit bearing in the camera frame.
In Fig.~\ref*{fig:spherical-projection}, the $z'$ axis aligns with $\mathbf d_g$, while $x'$ and $y'$ span the tangent plane to the unit sphere at $\mathbf d_g$.
Let the rows of $\mathbf B_g\in\mathbb R^{2\times3}$ form an orthonormal basis for this plane, so that $\mathbf B_g^\mathsf T\mathbf B_g=\mathbf I-\mathbf d_g\mathbf d_g^\mathsf T$.
Under first-order projection onto the unit sphere, the tangent-plane covariance and the Jacobian mapping tangent-plane coordinates to ERP pixels are
\begin{equation}
  \boldsymbol\Sigma'_g=\frac{\mathbf B_g\boldsymbol\Sigma_g^C\mathbf B_g^\mathsf T}{R_g^2},
  \qquad \mathbf A_g=R_g\mathbf J_{\pi,g}\mathbf B_g^\mathsf T,
  \label{eq:tangent-covariance}
\end{equation}
where $\boldsymbol\Sigma_g^C$ is the camera-frame covariance and $\mathbf J_{\pi,g}$ is the projection Jacobian evaluated at the Gaussian center.
Since ERP projection is invariant to radial scaling, $\mathbf J_{\pi,g}\mathbf d_g=\mathbf 0$ and hence $\mathbf J_{\pi,g}\mathbf B_g^\mathsf T\mathbf B_g=\mathbf J_{\pi,g}$.
Substituting Eq.~\eqref{eq:tangent-covariance} and adding the low-pass term gives
\begin{equation}
  \begin{aligned}
    \boldsymbol\Sigma^{\mathrm{img}}_g
      &=\mathbf A_g\boldsymbol\Sigma'_g\mathbf A_g^\mathsf T+\epsilon\mathbf I\\
      &=\mathbf J_{\pi,g}\boldsymbol\Sigma_g^C\mathbf J_{\pi,g}^\mathsf T+\epsilon\mathbf I.
  \end{aligned}
  \label{eq:supp-covariance-equivalence}
\end{equation}

\section{Adaptive Voxel-Scale Selection}
\label{sec:supp-adaptive-voxel}

Three LiDAR voxel maps with base voxel sizes of 0.5, 1, and 2 m are updated,
 each retaining its octree subdivision.
The active scale is initialized to 0.5 m.
The scales are periodically evaluated on a shared subset of scan points using two-fold cross-validation.
Pose increments estimated on one subset are evaluated on the other, with their roles exchanged.
We score each scale using the negative log-likelihood of the held-out point-to-plane residuals, accounting for measurement and plane uncertainty.
Unmatched points incur a penalty.
Scores are averaged over recent evaluations.
We switch scales only if a candidate consistently yields a better score and retains at least 80\% of the current scale's posterior pose information in every direction.

\begin{table}[!htbp]
\centering
\small
\caption{\textbf{Voxel-scale ablation on all 20 WHOLE sequences.} Values are translational endpoint drift (m, $\downarrow$).}
\label{tab:supp-whole-voxel}
\begin{tabularx}{\columnwidth}{@{}lYY@{}}
\toprule
Sequence & Fixed 0.5 m & Adaptive \\
\midrule
corridor\_01 & 0.272 & 0.272 \\
garage\_01 & 4.007 & 4.007 \\
garage\_02 & 1.350 & 1.350 \\
garage\_03 & 0.040 & 0.040 \\
garage\_04 & 0.015 & 0.015 \\
indoor\_01 & 0.020 & 0.020 \\
indoor\_02 & 0.008 & 0.008 \\
motion\_blur\_01 & 0.001 & 0.001 \\
motion\_blur\_02 & 0.004 & 0.004 \\
motion\_blur\_03 & \textbf{0.005} & 0.006 \\
motion\_blur\_04 & 0.005 & 0.005 \\
night\_01 & 0.004 & 0.004 \\
night\_02 & 0.009 & 0.009 \\
night\_03 & 6.418 & \textbf{4.983} \\
night\_04 & 0.004 & 0.004 \\
outdoor\_01 & 0.886 & \textbf{0.057} \\
outdoor\_02 & \textbf{0.008} & 0.009 \\
outdoor\_03 & 0.016 & 0.016 \\
outdoor\_04 & 6.369 & \textbf{5.815} \\
outdoor\_05 & 10.292 & \textbf{3.334} \\
\midrule
Mean & 1.487 & \textbf{0.998} \\
\bottomrule
\end{tabularx}
\end{table}

\textbf{Ablation on WHOLE.}
Adaptive selection reduces mean endpoint drift from 1.487 to 0.998\,m (32.9\%) relative to the fixed-scale baseline (Table~\ref{tab:supp-whole-voxel}).
On 16 sequences, the frontend retains the initial scale for most of the run, with less than 1\,mm change in endpoint drift.
All four clear improvements occur in outdoor sequences, where the frontend uses coarser scales for most of each run.
When active, the coarser maps match a larger share of the same sampled LiDAR points on average than the map at the initial scale.
In open scenes, larger voxels can group more LiDAR points from sparsely sampled distant surfaces to fit local planes.

\section{Bearing-Ray Triangulation}
\label{sec:supp-triangulation}
For the blind-zone constraints in Sec.~\ref*{sec:panoramic-odometry}, let $\mathcal H_i$ index the frames observing feature $i$ before the current frame $k$.
For each $j\in\mathcal H_i$, the world-frame unit bearing is $\mathbf d_{ij}^W=\mathbf R_{C_j}^W\mathbf d(u_{ij},v_{ij})$, and the camera center is $\mathbf c_j^W$.
The triangulated point minimizes the sum of squared distances to these bearing rays:
\begin{equation}
  \mathbf P_i^W=\underset{\mathbf P}{\arg\min}
  \sum_{j\in\mathcal H_i}\left\|
  (\mathbf I-\mathbf d_{ij}^W\mathbf d_{ij}^{W\mathsf T})
  (\mathbf P-\mathbf c_j^W)\right\|_2^2.
  \label{eq:blind-triangulation}
\end{equation}
Each matrix $\mathbf I-\mathbf d_{ij}^W\mathbf d_{ij}^{W\mathsf T}$ projects the camera-to-point displacement onto the plane perpendicular to the observed ray.
With sufficient parallax, we retain points with positive depths and directions consistent with the historical bearings to form the current-frame residual in Eq.~(\ref*{eq:blind-bearing-residual}).

\textbf{Uncertainty Propagation.}
To weight the current-frame bearing residual, we first estimate the covariance of the triangulated point and then propagate it to the residual.
With $n_i=|\mathcal H_i|$ historical observations and $\mathbf A_i=\sum_{j\in\mathcal H_i}(\mathbf I-\mathbf d_{ij}^W\mathbf d_{ij}^{W\mathsf T})$ summarizing their bearing geometry, we approximate the point and residual covariances as
\begin{equation}
  \begin{aligned}
    \boldsymbol\Sigma_{P,i}
      &=\bigl(v_i+(n_i-1)v_i^{\mathrm{sh}}\bigr)\mathbf A_i^{-1},\\
    \boldsymbol\Sigma_{r,i}
      &=\mathbf J_i\boldsymbol\Sigma_{P,i}\mathbf J_i^\mathsf T
        +\sigma_\theta^2\mathbf I_2.
  \end{aligned}
  \label{eq:supp-triangulation-uncertainty}
\end{equation}
For each historical ray, we estimate the variance perpendicular to the ray from range-scaled angular noise and pose uncertainty.
$v_i$ is the largest of these variances, and $v_i^{\mathrm{sh}}$ is the largest shared pose contribution across the historical views.
The shared term accounts for pose uncertainty common to historical views, which would otherwise be averaged down as independent noise.
Here, $\mathbf J_i=\partial\mathbf r_{ik}/\partial\mathbf P_i^W$ includes the world-to-camera rotation, and $\sigma_\theta$ is the standard deviation of angular observation noise in the current frame.

\section{Patch Brightness Uncertainty}
\label{sec:supp-patch-brightness}

For a patch with $N$ pixels, let $\mathbf i_{\mathrm{ref}},\mathbf i_{\mathrm{cur}}\in\mathbb R^N$ contain the reference and geometrically aligned current intensities.
The aligned patch intensities are modeled as
\begin{equation}
  \mathbf i_{\mathrm{cur}}=(1+\alpha)\mathbf i_{\mathrm{ref}}+b\mathbf 1+\mathbf n,
  \label{eq:supp-patch-brightness-model}
\end{equation}
where $\alpha$ is a relative gain perturbation, $b$ is an additive offset, and $\mathbf n$ is pixel noise.
The gain, offset, and pixel noise are assumed mutually independent and zero-mean, with $\mathbb E[\alpha^2]=\sigma_\alpha^2$, $\mathbb E[b^2]=\sigma_b^2$, and $\mathbb E[\mathbf n\mathbf n^\mathsf T]=\sigma_n^2\mathbf I$.
Conditioned on the reference patch, the residual $\mathbf r_{\mathrm{photo}}=\mathbf i_{\mathrm{cur}}-\mathbf i_{\mathrm{ref}}$ has covariance
\begin{equation}
  \begin{aligned}
    \mathbf C_{\mathrm{photo}}
      &=\mathbb E[\mathbf r_{\mathrm{photo}}\mathbf r_{\mathrm{photo}}^\mathsf T]\\
      &=\sigma_n^2\mathbf I
        +\sigma_\alpha^2\mathbf i_{\mathrm{ref}}\mathbf i_{\mathrm{ref}}^\mathsf T
        +\sigma_b^2\mathbf 1\mathbf 1^\mathsf T,
  \end{aligned}
  \label{eq:supp-patch-brightness-covariance}
\end{equation}
where the cross terms vanish by independence and zero means, yielding Eq.~(\ref*{eq:patch-brightness-covariance}).

\section{Panoramic Depth Estimator Ablation}
\label{sec:supp-depth-estimator}

\begin{figure}[!htbp]
\centering
\includegraphics[width=0.79\linewidth]{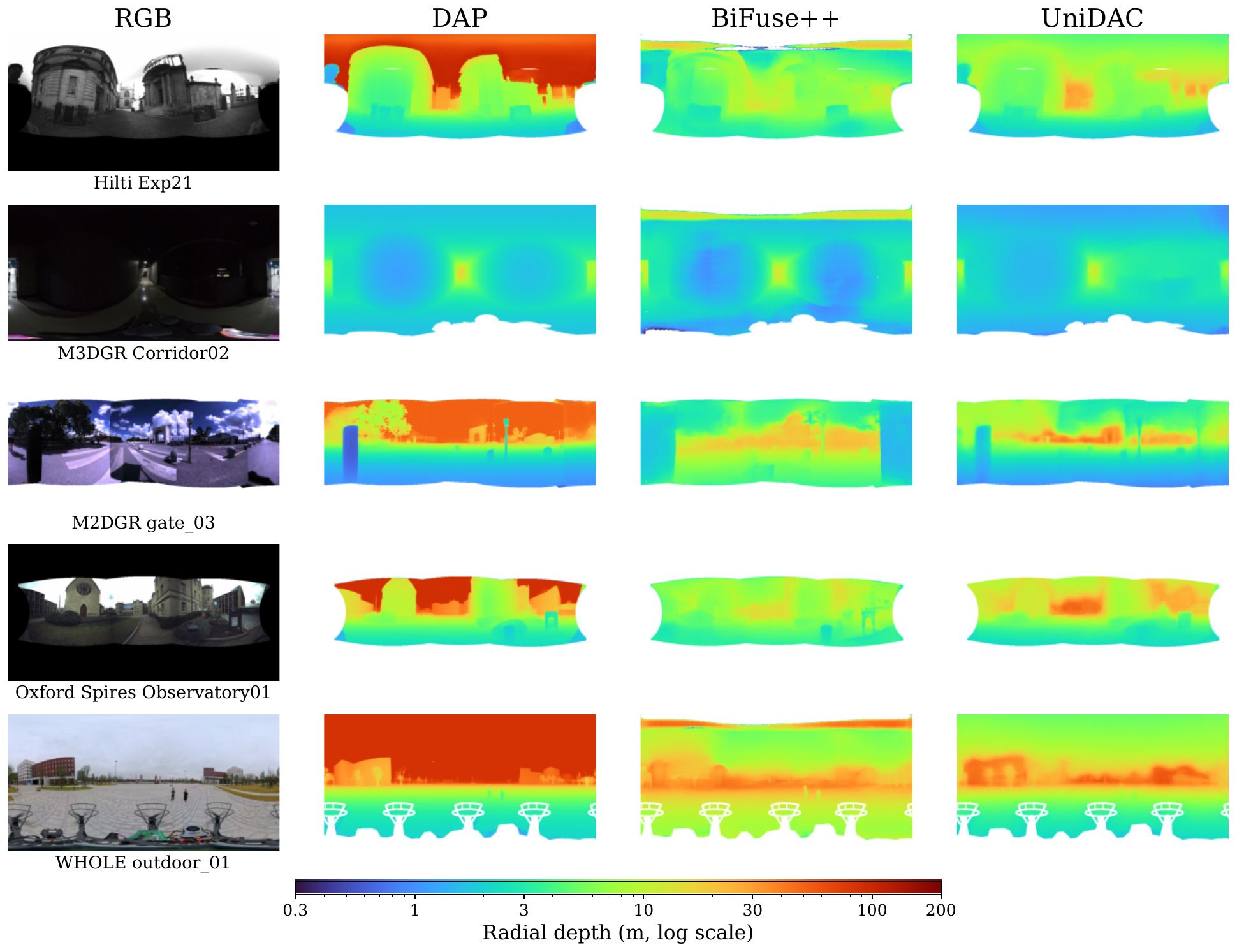}
\caption{LiDAR-aligned panoramic depth predictions from DAP, BiFuse++, and UniDAC.}
\label{fig:supp-depth-estimator}
\end{figure}

We compare DAP~\cite{dap}, BiFuse++~\cite{bifusepp}, and UniDAC~\cite{ganesan2026unidac} using public pretrained weights without fine-tuning.
DAP achieves the lowest Render-Depth-L1 on four of the five sequences and the highest PSNR and SSIM on three (Table~\ref{tab:supp-depth-estimator}).
On gate\_03, UniDAC has the lowest raw depth error, while DAP has the lowest rendered error after LiDAR alignment and depth supervision.
Figure~\ref{fig:supp-depth-estimator} shows that DAP generally produces clearer depth boundaries between foreground structures and the background.
These reconstruction results favor DAP as the default estimator, while BiFuse++ has the shortest runtime and lowest peak VRAM usage on all five sequences.

\begin{table}[!htbp]
\begingroup
\centering
\small
\setlength{\tabcolsep}{5pt}
\caption{\textbf{Panoramic depth estimator ablation.}
Raw-Depth-L1 measures prediction error before LiDAR scale alignment; Render-Depth-L1 measures final map error.
Rendering metrics use in-sequence novel ERP views.}
\label{tab:supp-depth-estimator}
\begin{tabularx}{\textwidth}{@{}ll*{7}{Y}@{}}
\toprule
Sequence & Estimator & \shortstack{PSNR\\(dB) $\uparrow$} & \shortstack{SSIM\\$\uparrow$}
& \shortstack{Raw\\Depth-L1\\{[m]} $\downarrow$} & \shortstack{Render\\Depth-L1\\{[m]} $\downarrow$} & \shortstack{$T_{\rm total}$\\{[s]} $\downarrow$} & \shortstack{VRAM\\{[GiB]} $\downarrow$} & \shortstack{RAM\\{[GiB]} $\downarrow$} \\
\midrule
Exp21 & DAP & \textbf{23.52} & \textbf{0.910} & \textbf{3.053} & \textbf{0.531} & 270.4 & 5.46 & 16.70 \\
 & BiFuse++ & 22.68 & 0.892 & 6.949 & 0.560 & \textbf{184.3} & \textbf{2.96} & \textbf{16.40} \\
 & UniDAC & 22.64 & 0.894 & 4.792 & 0.544 & 257.1 & 7.27 & 17.32 \\
\midrule
Corridor02 & DAP & 25.79 & 0.879 & \textbf{0.558} & \textbf{0.221} & 569.0 & 7.48 & 9.59 \\
 & BiFuse++ & 25.85 & 0.879 & 0.964 & 0.227 & \textbf{532.5} & \textbf{5.02} & \textbf{9.51} \\
 & UniDAC & \textbf{25.87} & \textbf{0.881} & 0.587 & 0.230 & 567.7 & 8.91 & 9.89 \\
\midrule
gate\_03 & DAP & \textbf{16.75} & \textbf{0.595} & 15.037 & \textbf{2.022} & 374.0 & 5.88 & 17.34 \\
 & BiFuse++ & 16.71 & 0.580 & 12.970 & 2.098 & \textbf{296.2} & \textbf{3.34} & \textbf{14.35} \\
 & UniDAC & 16.45 & 0.572 & \textbf{9.406} & 2.100 & 320.1 & 7.02 & 14.67 \\
\midrule
Observatory01 & DAP & \textbf{25.11} & \textbf{0.822} & \textbf{2.429} & \textbf{0.482} & 482.7 & 7.63 & 12.56 \\
 & BiFuse++ & 24.80 & 0.814 & 4.985 & 0.525 & \textbf{390.4} & \textbf{4.37} & 11.96 \\
 & UniDAC & 24.93 & 0.814 & 8.432 & 0.516 & 420.1 & 8.21 & \textbf{11.59} \\
\midrule
outdoor\_01 & DAP & 24.94 & 0.858 & 18.743 & 1.435 & 250.5 & 5.46 & \textbf{11.27} \\
 & BiFuse++ & 24.65 & 0.832 & 12.945 & 1.454 & \textbf{194.1} & \textbf{3.00} & 11.44 \\
 & UniDAC & \textbf{24.95} & \textbf{0.859} & \textbf{7.031} & \textbf{1.413} & 269.8 & 7.21 & 11.66 \\
\bottomrule
\end{tabularx}
\par
\endgroup
\end{table}

\FloatBarrier
\section{LiDAR Sampling Density Ablation}
\label{sec:supp-lidar-sampling}

\begin{figure}[!htbp]
\centering
\includegraphics[width=0.63\linewidth]{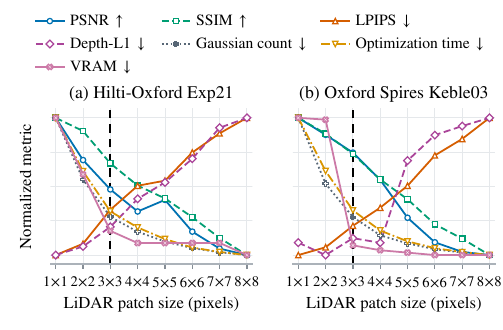}
\caption{\textbf{LiDAR patch-size ablation.}
Each curve is min--max normalized.
Vertical dashed lines mark the default $3\times3$ setting.}
\label{fig:supp-lidar-sampling}
\end{figure}

To initialize Gaussians from LiDAR, we retain the valid point with the smallest radial depth in each ERP patch.
We vary the patch size from $1\times1$ to $8\times8$ pixels on the first 600 input frames of Hilti-Oxford Exp21 and Oxford Spires Keble03, with DAP sampling fixed.
Compared with $1\times1$ patches, $3\times3$ patches reduce the final Gaussian count by over 60\% on both sequences, while lowering optimization time and peak VRAM usage.
These savings come with a PSNR drop below 0.3\,dB and a Depth-L1 increase below 0.02\,m.
Increasing the patch size beyond $3\times3$ saves more optimization time but lowers SSIM and raises LPIPS (Fig.~\ref{fig:supp-lidar-sampling}).

\section{Keyframe Selection Ablation}
\label{sec:supp-keyframe-selection}

On the first 1,000 input frames of each sequence, we compare fixed-interval sampling (Fixed), co-visibility and depth-normalized translation (C+T)~\cite{monogs}, spherical angular displacement (Parallax), and translation distance (Displacement).
Fixed achieves the lowest LPIPS on four of the five sequences (Table~\ref{tab:supp-keyframe-selection}).
On Observatory01, Parallax selects 351 keyframes versus 198 for Fixed, improving PSNR by 0.43\,dB at more than twice the optimization time.
We use Fixed as the default to keep the keyframe rate predictable while retaining competitive reconstruction quality.

\begingroup
\small
\setlength{\tabcolsep}{3pt}
\renewcommand{\arraystretch}{1.06}
\setlength{\LTcapwidth}{\linewidth}
\begin{longtable}{@{}l*{5}{>{\centering\arraybackslash}p{.155\linewidth}}@{}}
\caption{\textbf{Keyframe selection ablation on shared held-out ERP views.}}
\label{tab:supp-keyframe-selection}\\
\toprule
Policy & \shortstack{PSNR$\uparrow$\\(dB)} & SSIM$\uparrow$ & LPIPS$\downarrow$ & \shortstack{Depth-L1\\(m) $\downarrow$} & \shortstack{Opt. time\\(s) $\downarrow$} \\
\midrule
\endfirsthead
\multicolumn{6}{c}{\tablename~\thetable\ (continued)}\\
\toprule
Policy & \shortstack{PSNR$\uparrow$\\(dB)} & SSIM$\uparrow$ & LPIPS$\downarrow$ & \shortstack{Depth-L1\\(m) $\downarrow$} & \shortstack{Opt. time\\(s) $\downarrow$} \\
\midrule
\endhead
\bottomrule
\endfoot
\multicolumn{6}{@{}l}{\textbf{Hilti-Oxford Exp21}} \\*
Fixed & 23.51 & \textbf{0.916} & \textbf{0.125} & \textbf{0.481} & 203.2 \\*
C+T & 23.25 & 0.910 & 0.136 & 0.512 & \textbf{125.7} \\*
Parallax & \textbf{23.70} & 0.915 & 0.134 & 0.494 & 268.0 \\*
Displacement & 23.39 & 0.913 & 0.135 & 0.499 & 163.5 \\
\midrule
\multicolumn{6}{@{}l}{\textbf{M3DGR Corridor02}} \\*
Fixed & 25.95 & 0.890 & \textbf{0.160} & 0.220 & 507.9 \\*
C+T & \textbf{26.37} & \textbf{0.896} & 0.172 & \textbf{0.215} & 1158.0 \\*
Parallax & 25.61 & 0.883 & 0.173 & 0.236 & 562.4 \\*
Displacement & 25.66 & 0.884 & 0.161 & 0.234 & \textbf{334.6} \\
\midrule
\multicolumn{6}{@{}l}{\textbf{M2DGR gate\_03}} \\*
Fixed & 16.80 & \textbf{0.607} & \textbf{0.359} & \textbf{1.982} & 298.3 \\*
C+T & \textbf{16.83} & 0.605 & 0.362 & 2.077 & 339.7 \\*
Parallax & 16.35 & 0.588 & 0.377 & 2.128 & 268.7 \\*
Displacement & 16.52 & 0.596 & 0.372 & 2.078 & \textbf{211.4} \\
\midrule
\multicolumn{6}{@{}l}{\textbf{Oxford Spires Observatory01}} \\*
Fixed & 25.09 & 0.821 & 0.120 & 0.497 & \textbf{411.7} \\*
C+T & 25.44 & 0.830 & \textbf{0.113} & 0.485 & 744.3 \\*
Parallax & \textbf{25.52} & \textbf{0.835} & 0.115 & \textbf{0.478} & 965.9 \\*
Displacement & 25.35 & 0.830 & 0.114 & 0.491 & 585.4 \\
\midrule
\multicolumn{6}{@{}l}{\textbf{WHOLE outdoor\_01}} \\*
Fixed & 25.25 & 0.860 & \textbf{0.128} & 1.361 & 200.1 \\*
C+T & 24.88 & 0.846 & 0.136 & 1.483 & \textbf{113.4} \\*
Parallax & \textbf{25.31} & \textbf{0.864} & 0.131 & \textbf{1.349} & 371.9 \\*
Displacement & 25.05 & 0.858 & 0.133 & 1.410 & 238.5 \\
\end{longtable}
\endgroup

\section{Optimization View Budget Ablation}
\label{sec:supp-optimization-budget}

\begin{figure}[!htbp]
\centering
\includegraphics[width=0.57\linewidth]{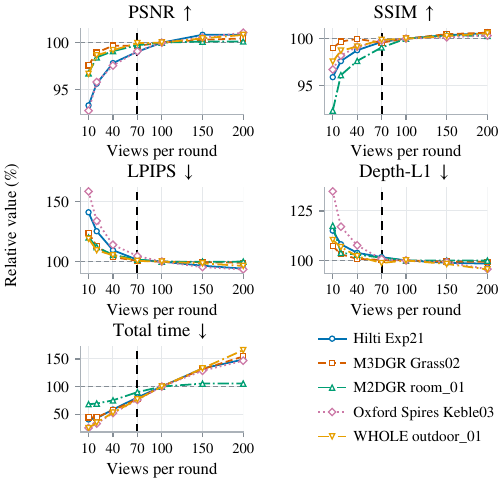}
\caption{\textbf{Optimization view budget ablation.} Each metric is normalized to its value at 100 views per round.}
\label{fig:supp-optimization-budget}
\end{figure}

Across five sequences, reducing the budget from $100$ to $70$ views per round lowers total online runtime by 20.1\% on average, while PSNR decreases by 0.01--0.25 dB and Depth-L1 changes by $-0.026$ to $0.011$ m (Fig.~\ref{fig:supp-optimization-budget}).
These results support a budget of $70$ views per round as a balance between reconstruction quality and efficiency.

\clearpage
\section{Per-Sequence Evaluation}
\label{sec:supp-sequence-evaluation}

Tables~\ref{tab:supp-sequence-odometry} and~\ref{tab:supp-sequence-localization} report per-sequence localization results corresponding to Tables~\ref*{tab:odometry} and~\ref*{tab:sparse-trajectory}, respectively.
Tables~\ref{tab:supp-sequence-rendering} and~\ref{tab:supp-sequence-geometry} provide the rendering and geometry results underlying Table~\ref*{tab:front-rendering}.

\begingroup
\fontsize{9}{10.6}\selectfont
\setlength{\tabcolsep}{2pt}
\setlength{\LTcapwidth}{\textwidth}
\begin{longtable}{@{}l*{7}{c}@{}}
\caption{\textbf{Per-sequence localization accuracy.} ATE RMSE (m, $\downarrow$).
OOM denotes GPU out-of-memory termination. \textcolor{red}{$*$} indicates failure. }
\label{tab:supp-sequence-odometry}\\
\toprule
\shortstack{Sequence\\(image counts)} & FAST-LIO2 & FAST-LIVO2 & \shortstack{Gaussian-\\LIC2} & \shortstack{Ground-Fusion++\\(LIC)} & \shortstack{ODGS-\\SLAM} & \shortstack{PanoGS-\\SLAM} & \shortstack{\textbf{GOLD-SLAM}\\\textbf{(Ours)}} \\
\midrule
\endfirsthead
\multicolumn{8}{c}{\tablename~\thetable\ (continued)}\\
\toprule
\shortstack{Sequence\\(image counts)} & FAST-LIO2 & FAST-LIVO2 & \shortstack{Gaussian-\\LIC2} & \shortstack{Ground-Fusion++\\(LIC)} & \shortstack{ODGS-\\SLAM} & \shortstack{PanoGS-\\SLAM} & \shortstack{\textbf{GOLD-SLAM}\\\textbf{(Ours)}} \\
\midrule
\endhead
\midrule
\multicolumn{8}{r}{Continued on next page}\\
\endfoot
\bottomrule
\endlastfoot
\multicolumn{8}{l}{\textbf{Hilti-Oxford 2022}} \\*
Exp04 (1257) & \rankbest{0.053} & \rankbest{0.053} & \rankbest{0.053} & \rankbest{0.053} & 7.345 & 7.408 & \ranksecond{0.054} \\
Exp14 (740) & 0.318 & \ranksecond{0.039} & 0.311 & 0.367 & 7.653 & 7.662 & \rankbest{0.037} \\
Exp18 (1093) & 6.959 & \ranksecond{0.112} & 0.305 & 30.296$^{\textcolor{red}{*}}$ & 6.229 & 5.990 & \rankbest{0.100} \\
Exp21 (1527) & 0.132 & \ranksecond{0.051} & 0.115 & 0.054 & 15.432 & 15.606 & \rankbest{0.048} \\
\midrule
\multicolumn{8}{l}{\textbf{M3DGR}} \\*
Grass01 (4318) & \rankbest{0.530} & 0.533 & \ranksecond{0.531} & 0.545 & 4.787 & 20.180$^{\textcolor{red}{*}}$ & 0.532 \\
Grass02 (2594) & \rankbest{0.428} & \rankbest{0.428} & \ranksecond{0.429} & 0.435 & 3.764 & 5.930 & \rankbest{0.428} \\
Varying-illu05 (4000) & 0.096 & \ranksecond{0.078} & 0.177 & 0.160 & 1.141 & 61.840$^{\textcolor{red}{*}}$ & \rankbest{0.073} \\
\midrule
\multicolumn{8}{l}{\textbf{M2DGR}} \\*
room\_01 (724) & 0.400 & \rankbest{0.393} & 0.398 & \ranksecond{0.395} & 0.441 & 0.413 & \rankbest{0.393} \\
door\_02 (1256) & 0.340 & 0.330 & 0.340 & \rankbest{0.316} & 4.721 & 10.280 & \ranksecond{0.328} \\
hall\_02 (1279) & \rankbest{0.501} & \rankbest{0.501} & \ranksecond{0.502} & 0.505 & 1.786 & 2.399 & \rankbest{0.501} \\
gate\_03 (2771) & \ranksecond{0.190} & 0.217 & \rankbest{0.180} & \rankbest{0.180} & OOM & 37.414$^{\textcolor{red}{*}}$ & 0.218 \\
\midrule
\multicolumn{8}{l}{\textbf{Oxford Spires}} \\*
Keble03 (5601) & 0.071 & 0.097 & 0.245 & \rankbest{0.043} & OOM & 31.541$^{\textcolor{red}{*}}$ & \ranksecond{0.059} \\
Observatory01 (5632) & \rankbest{0.064} & 0.122 & 0.632 & 0.269 & 5.014 & 28.133$^{\textcolor{red}{*}}$ & \ranksecond{0.086} \\
ChristChurch03 (6077) & 0.027 & \ranksecond{0.023} & 0.757 & 0.037 & 5.705 & 16.230 & \rankbest{0.014} \\
Blenheim01 (7894) & \rankbest{0.139} & 0.158 & 0.618 & 0.997 & OOM & 55.477$^{\textcolor{red}{*}}$ & \ranksecond{0.150} \\
\end{longtable}
\endgroup

\begingroup
\fontsize{9}{10.6}\selectfont
\setlength{\tabcolsep}{2pt}
\setlength{\LTcapwidth}{\textwidth}
\renewcommand{\arraystretch}{1.02}
\begin{longtable}{@{}l*{7}{c}@{}}
\caption{\textbf{Per-sequence translational endpoint drift (m, $\downarrow$).} }
\label{tab:supp-sequence-localization}\\
\toprule
\shortstack[l]{Sequence\\(image counts)} & FAST-LIO2 & FAST-LIVO2 & \shortstack{Gaussian-\\LIC2} & \shortstack{Ground-Fusion++\\(LIC)} & \shortstack{ODGS-\\SLAM} & \shortstack{PanoGS-\\SLAM} & \shortstack{GOLD-SLAM\\(Ours)} \\
\midrule
\endfirsthead
\multicolumn{8}{c}{\tablename~\thetable\ (continued)}\\
\toprule
\shortstack[l]{Sequence\\(image counts)} & FAST-LIO2 & FAST-LIVO2 & \shortstack{Gaussian-\\LIC2} & \shortstack{Ground-Fusion++\\(LIC)} & \shortstack{ODGS-\\SLAM} & \shortstack{PanoGS-\\SLAM} & \shortstack{GOLD-SLAM\\(Ours)} \\
\midrule
\endhead
\midrule
\multicolumn{8}{r}{Continued on next page}\\
\endfoot
\bottomrule
\endlastfoot
\multicolumn{8}{@{}l}{\textbf{M3DGR}} \\*
Corridor01 (6058) & 18.317 & 34.548$^{\textcolor{red}{*}}$ & 14.272 & \ranksecond{2.700} & 11.309 & 49.698$^{\textcolor{red}{*}}$ & \rankbest{1.161} \\[1pt]
Corridor02 (4400) & 17.491 & \ranksecond{1.319} & 19.512 & 5.065 & 9.252 & 10.223 & \rankbest{0.274} \\[1pt]
\midrule
\multicolumn{8}{@{}l}{\textbf{WHOLE}} \\*
corridor\_01 (3395) & 13.602 & 0.273 & \rankbest{0.011} & 1225.608$^{\textcolor{red}{*}}$ & 18.978 & 7.863 & \ranksecond{0.272} \\[1pt]
garage\_01 (2900) & \rankbest{0.055} & 3.026 & 4.136 & \ranksecond{2.150} & 23.219$^{\textcolor{red}{*}}$ & 21.788$^{\textcolor{red}{*}}$ & 4.007 \\[1pt]
garage\_02 (4712) & \rankbest{0.719} & 1.381 & 1.964 & \ranksecond{1.005} & 106.074$^{\textcolor{red}{*}}$ & 7.434 & 1.350 \\[1pt]
garage\_03 (2330) & 0.086 & \rankbest{0.035} & 0.191 & 0.723 & 2.123 & 9.815 & \ranksecond{0.040} \\[1pt]
garage\_04 (2810) & 0.042 & \ranksecond{0.011} & 0.022 & \rankbest{0.008} & 4.673 & 22.782$^{\textcolor{red}{*}}$ & 0.015 \\[1pt]
indoor\_01 (1765) & 0.040 & 0.023 & 0.021 & \rankbest{0.004} & 2.655 & 5.081 & \ranksecond{0.020} \\[1pt]
indoor\_02 (1528) & 0.034 & \ranksecond{0.005} & 0.018 & \rankbest{0.002} & 9.327 & 3.251 & 0.008 \\[1pt]
motion\_blur\_01 (1916) & 0.024 & \ranksecond{0.006} & 0.014 & \ranksecond{0.006} & 32.252$^{\textcolor{red}{*}}$ & 8.572 & \rankbest{0.001} \\[1pt]
motion\_blur\_02 (1547) & 0.019 & 0.008 & 0.019 & \ranksecond{0.005} & 21.534$^{\textcolor{red}{*}}$ & 2.622 & \rankbest{0.004} \\[1pt]
motion\_blur\_03 (596) & 0.010 & \rankbest{0.003} & 0.011 & \rankbest{0.003} & 3.196 & 2.354 & \ranksecond{0.006} \\[1pt]
motion\_blur\_04 (1409) & 0.025 & \ranksecond{0.006} & 0.033 & 226.881$^{\textcolor{red}{*}}$ & 8.056 & 2.821 & \rankbest{0.005} \\[1pt]
night\_01 (1602) & 0.048 & \ranksecond{0.004} & 0.052 & \rankbest{0.003} & 8.786 & 2.435 & \ranksecond{0.004} \\[1pt]
night\_02 (2474) & 0.038 & \ranksecond{0.010} & 0.050 & \rankbest{0.009} & 1.579 & 2.087 & \rankbest{0.009} \\[1pt]
night\_03 (3843) & 7.300 & \ranksecond{6.873} & 7.266 & 9.729 & 47.383$^{\textcolor{red}{*}}$ & 59.671$^{\textcolor{red}{*}}$ & \rankbest{4.983} \\[1pt]
night\_04 (1734) & 0.021 & \ranksecond{0.005} & 0.014 & 0.394 & 29.028$^{\textcolor{red}{*}}$ & 4.365 & \rankbest{0.004} \\[1pt]
outdoor\_01 (2881) & \ranksecond{0.399} & 0.888 & 1.473 & 0.906 & 24.638$^{\textcolor{red}{*}}$ & 0.994 & \rankbest{0.057} \\[1pt]
outdoor\_02 (1138) & 0.025 & \rankbest{0.008} & 0.021 & \ranksecond{0.009} & 5.721 & 3.093 & \ranksecond{0.009} \\[1pt]
outdoor\_03 (1663) & 0.028 & 0.020 & 0.030 & \ranksecond{0.019} & 3.541 & 9.256 & \rankbest{0.016} \\[1pt]
outdoor\_04 (4146) & 8.474 & \ranksecond{6.868} & 7.327 & 8.005 & 56.730$^{\textcolor{red}{*}}$ & 22.207$^{\textcolor{red}{*}}$ & \rankbest{5.815} \\[1pt]
outdoor\_05 (3578) & 11.829 & \ranksecond{10.544} & 11.323 & 10.978 & 67.498$^{\textcolor{red}{*}}$ & 26.106$^{\textcolor{red}{*}}$ & \rankbest{3.334} \\[1pt]
\end{longtable}
\endgroup

\clearpage
\begingroup
\fontsize{9}{10.6}\selectfont
\setlength{\tabcolsep}{2.1pt}
\renewcommand{\arraystretch}{1.0}
\setlength{\LTleft}{\fill}
\setlength{\LTright}{\fill}
\setlength{\LTcapwidth}{\linewidth}
\begin{longtable}{@{}l*{4}{@{\hspace{8pt}}rrr}@{}}
\caption{\textbf{Per-sequence novel-view rendering evaluation.} PSNR is in dB.}
\label{tab:supp-sequence-rendering}\\
\toprule
\shortstack[l]{Sequence\\(image counts)} & \multicolumn{3}{c}{Gaussian-LIC2} & \multicolumn{3}{c}{ODGS-SLAM} & \multicolumn{3}{c}{PanoGS-SLAM} & \multicolumn{3}{c}{\textbf{GOLD-SLAM}} \\
 & PSNR$\uparrow$ & SSIM$\uparrow$ & LPIPS$\downarrow$ & PSNR$\uparrow$ & SSIM$\uparrow$ & LPIPS$\downarrow$ & PSNR$\uparrow$ & SSIM$\uparrow$ & LPIPS$\downarrow$ & PSNR$\uparrow$ & SSIM$\uparrow$ & LPIPS$\downarrow$ \\
\midrule
\endfirsthead
\multicolumn{13}{c}{\tablename~\thetable\ (continued)}\\
\toprule
\shortstack[l]{Sequence\\(image counts)} & \multicolumn{3}{c}{Gaussian-LIC2} & \multicolumn{3}{c}{ODGS-SLAM} & \multicolumn{3}{c}{PanoGS-SLAM} & \multicolumn{3}{c}{\textbf{GOLD-SLAM}} \\
 & PSNR$\uparrow$ & SSIM$\uparrow$ & LPIPS$\downarrow$ & PSNR$\uparrow$ & SSIM$\uparrow$ & LPIPS$\downarrow$ & PSNR$\uparrow$ & SSIM$\uparrow$ & LPIPS$\downarrow$ & PSNR$\uparrow$ & SSIM$\uparrow$ & LPIPS$\downarrow$ \\
\midrule
\endhead
\midrule
\multicolumn{13}{r}{Continued on next page}\\
\endfoot
\bottomrule
\endlastfoot
\multicolumn{13}{@{}l}{\textbf{Hilti-Oxford 2022}}\\*
\addlinespace[2pt]
Exp04 (1257) & \rankbest{26.39} & \ranksecond{0.895} & \ranksecond{0.320} & 15.80 & 0.552 & 0.582 & 19.51 & 0.800 & 0.571 & \ranksecond{26.36} & \rankbest{0.902} & \rankbest{0.283} \\[0.5pt]
Exp14 (740) & \ranksecond{27.78} & \ranksecond{0.905} & \ranksecond{0.199} & 14.54 & 0.619 & 0.615 & 19.45 & 0.820 & 0.539 & \rankbest{29.14} & \rankbest{0.938} & \rankbest{0.147} \\[0.5pt]
Exp18 (1093) & \ranksecond{23.61} & \ranksecond{0.856} & \ranksecond{0.258} & 13.49 & 0.438 & 0.590 & 17.18 & 0.705 & 0.538 & \rankbest{25.42} & \rankbest{0.888} & \rankbest{0.189} \\[0.5pt]
Exp21 (1527) & \ranksecond{27.46} & \ranksecond{0.922} & \ranksecond{0.165} & 11.71 & 0.365 & 0.652 & 20.33 & 0.779 & 0.462 & \rankbest{27.59} & \rankbest{0.935} & \rankbest{0.124} \\[0.5pt]
\midrule
\multicolumn{13}{@{}l}{\textbf{M3DGR}}\\*
\addlinespace[2pt]
Grass01 (4318) & \ranksecond{20.22} & \ranksecond{0.555} & \ranksecond{0.419} & 15.28 & 0.479 & 0.430 & \rankbest{20.62} & \rankbest{0.562} & 0.515 & 19.51 & 0.531 & \rankbest{0.417} \\[0.5pt]
Grass02 (2594) & \ranksecond{19.59} & \ranksecond{0.524} & \ranksecond{0.433} & 19.37 & 0.507 & \rankbest{0.378} & \rankbest{22.87} & \rankbest{0.626} & 0.436 & 18.92 & 0.509 & 0.440 \\[0.5pt]
Varying-illu05 (4000) & \rankbest{28.89} & \ranksecond{0.889} & \rankbest{0.269} & 23.53 & 0.683 & 0.319 & 25.38 & 0.866 & 0.424 & \ranksecond{28.49} & \rankbest{0.891} & \ranksecond{0.275} \\[0.5pt]
Corridor01 (6058) & \ranksecond{23.06} & \ranksecond{0.835} & \ranksecond{0.261} & 21.19 & 0.708 & 0.353 & 21.04 & 0.791 & 0.421 & \rankbest{25.14} & \rankbest{0.861} & \rankbest{0.209} \\[0.5pt]
Corridor02 (4400) & \ranksecond{22.06} & \ranksecond{0.827} & \ranksecond{0.279} & 20.41 & 0.726 & 0.338 & 21.98 & 0.813 & 0.395 & \rankbest{24.87} & \rankbest{0.861} & \rankbest{0.210} \\[0.5pt]
\midrule
\multicolumn{13}{@{}l}{\textbf{M2DGR}}\\*
\addlinespace[2pt]
room\_01 (724) & \rankbest{19.81} & \rankbest{0.639} & \rankbest{0.283} & 13.89 & 0.304 & 0.557 & 16.67 & 0.489 & 0.470 & \ranksecond{16.77} & \ranksecond{0.504} & \ranksecond{0.464} \\[0.5pt]
door\_02 (1256) & \ranksecond{19.38} & \ranksecond{0.715} & \ranksecond{0.353} & 14.04 & 0.426 & 0.521 & 15.94 & 0.586 & 0.540 & \rankbest{20.52} & \rankbest{0.759} & \rankbest{0.275} \\[0.5pt]
hall\_02 (1279) & \rankbest{17.39} & \rankbest{0.676} & \rankbest{0.348} & 11.69 & 0.366 & 0.631 & 14.69 & 0.582 & 0.498 & \ranksecond{16.56} & \ranksecond{0.661} & \ranksecond{0.367} \\[0.5pt]
gate\_03 (2771) & \ranksecond{19.19} & \ranksecond{0.703} & \ranksecond{0.385} & \multicolumn{3}{c}{OOM} & 16.50 & 0.621 & 0.528 & \rankbest{21.02} & \rankbest{0.755} & \rankbest{0.283} \\[0.5pt]
\midrule
\multicolumn{13}{@{}l}{\textbf{Oxford Spires}}\\*
\addlinespace[2pt]
Keble03 (5601) & \ranksecond{25.59} & \ranksecond{0.832} & \ranksecond{0.265} & \multicolumn{3}{c}{OOM} & 20.98 & 0.721 & 0.578 & \rankbest{27.94} & \rankbest{0.884} & \rankbest{0.145} \\[0.5pt]
Observatory01 (5632) & \ranksecond{23.50} & \ranksecond{0.809} & \ranksecond{0.288} & 11.50 & 0.462 & 0.634 & 20.60 & 0.710 & 0.587 & \rankbest{25.15} & \rankbest{0.858} & \rankbest{0.166} \\[0.5pt]
ChristChurch03 (6077) & \ranksecond{26.35} & \ranksecond{0.896} & \ranksecond{0.188} & 19.66 & 0.715 & 0.534 & 20.84 & 0.806 & 0.613 & \rankbest{26.96} & \rankbest{0.920} & \rankbest{0.133} \\[0.5pt]
Blenheim01 (7894) & \ranksecond{24.32} & \ranksecond{0.828} & \ranksecond{0.284} & \multicolumn{3}{c}{OOM} & 18.62 & 0.691 & 0.720 & \rankbest{24.55} & \rankbest{0.860} & \rankbest{0.212} \\[0.5pt]
\midrule
\multicolumn{13}{@{}l}{\textbf{WHOLE}}\\*
\addlinespace[2pt]
outdoor\_01 (2881) & \ranksecond{24.09} & \rankbest{0.823} & \ranksecond{0.223} & 15.60 & 0.551 & 0.451 & 20.51 & 0.700 & 0.426 & \rankbest{24.64} & \ranksecond{0.819} & \rankbest{0.167} \\[0.5pt]
indoor\_01 (1765) & \ranksecond{26.04} & \ranksecond{0.894} & \ranksecond{0.160} & 19.25 & 0.722 & 0.431 & 21.44 & 0.809 & 0.418 & \rankbest{27.96} & \rankbest{0.916} & \rankbest{0.116} \\[0.5pt]
indoor\_02 (1528) & \ranksecond{26.77} & \ranksecond{0.900} & \ranksecond{0.149} & 18.18 & 0.687 & 0.447 & 21.55 & 0.801 & 0.407 & \rankbest{28.12} & \rankbest{0.920} & \rankbest{0.108} \\[0.5pt]
motion\_blur\_01 (1916) & \ranksecond{24.70} & \ranksecond{0.868} & \ranksecond{0.192} & 15.76 & 0.558 & 0.579 & 18.14 & 0.763 & 0.538 & \rankbest{26.97} & \rankbest{0.904} & \rankbest{0.127} \\[0.5pt]
outdoor\_02 (1138) & \ranksecond{25.91} & \ranksecond{0.825} & \ranksecond{0.204} & 17.90 & 0.583 & 0.393 & 21.49 & 0.662 & 0.460 & \rankbest{27.43} & \rankbest{0.845} & \rankbest{0.161} \\[0.5pt]
outdoor\_03 (1663) & \ranksecond{26.59} & \ranksecond{0.858} & \ranksecond{0.173} & 17.19 & 0.598 & 0.366 & 23.02 & 0.733 & 0.348 & \rankbest{27.26} & \rankbest{0.874} & \rankbest{0.144} \\[0.5pt]
motion\_blur\_02 (1547) & \ranksecond{23.70} & \ranksecond{0.781} & \ranksecond{0.250} & 11.55 & 0.392 & 0.572 & 19.43 & 0.652 & 0.558 & \rankbest{24.73} & \rankbest{0.809} & \rankbest{0.192} \\[0.5pt]
motion\_blur\_03 (596) & \ranksecond{21.76} & \ranksecond{0.776} & \ranksecond{0.276} & 14.14 & 0.434 & 0.571 & 15.01 & 0.643 & 0.639 & \rankbest{25.44} & \rankbest{0.829} & \rankbest{0.182} \\[0.5pt]
night\_01 (1602) & \ranksecond{32.83} & \ranksecond{0.941} & \ranksecond{0.102} & 19.20 & 0.451 & 0.614 & 26.15 & 0.870 & 0.318 & \rankbest{33.38} & \rankbest{0.947} & \rankbest{0.090} \\[0.5pt]
night\_02 (2474) & \rankbest{34.14} & \ranksecond{0.951} & \ranksecond{0.097} & 21.68 & 0.516 & 0.464 & 27.88 & 0.891 & 0.297 & \ranksecond{34.13} & \rankbest{0.953} & \rankbest{0.096} \\[0.5pt]
corridor\_01 (3395) & \ranksecond{25.66} & \ranksecond{0.900} & \ranksecond{0.178} & 15.80 & 0.640 & 0.577 & 18.68 & 0.799 & 0.507 & \rankbest{26.80} & \rankbest{0.914} & \rankbest{0.131} \\[0.5pt]
motion\_blur\_04 (1409) & \ranksecond{22.17} & \ranksecond{0.821} & \ranksecond{0.253} & 12.16 & 0.444 & 0.678 & 16.78 & 0.730 & 0.590 & \rankbest{24.70} & \rankbest{0.857} & \rankbest{0.169} \\[0.5pt]
garage\_01 (2900) & \ranksecond{24.45} & \ranksecond{0.850} & \ranksecond{0.172} & 15.90 & 0.521 & 0.512 & 17.22 & 0.655 & 0.618 & \rankbest{25.09} & \rankbest{0.867} & \rankbest{0.138} \\[0.5pt]
garage\_02 (4712) & \rankbest{26.09} & \ranksecond{0.883} & \ranksecond{0.132} & 15.20 & 0.538 & 0.534 & 16.59 & 0.668 & 0.650 & \ranksecond{25.82} & \rankbest{0.888} & \rankbest{0.121} \\[0.5pt]
garage\_03 (2330) & \ranksecond{24.94} & \ranksecond{0.875} & \ranksecond{0.148} & 18.37 & 0.656 & 0.381 & 21.46 & 0.792 & 0.352 & \rankbest{25.07} & \rankbest{0.887} & \rankbest{0.135} \\[0.5pt]
garage\_04 (2810) & \rankbest{26.54} & \rankbest{0.892} & \rankbest{0.113} & 16.58 & 0.561 & 0.496 & 19.06 & 0.724 & 0.465 & \ranksecond{25.57} & \ranksecond{0.887} & \ranksecond{0.118} \\[0.5pt]
outdoor\_04 (4146) & \ranksecond{23.10} & \ranksecond{0.794} & \ranksecond{0.267} & 16.05 & 0.578 & 0.534 & 18.64 & 0.650 & 0.582 & \rankbest{23.74} & \rankbest{0.802} & \rankbest{0.224} \\[0.5pt]
outdoor\_05 (3578) & \ranksecond{23.04} & \ranksecond{0.786} & \ranksecond{0.279} & 15.75 & 0.552 & 0.506 & 17.93 & 0.624 & 0.615 & \rankbest{23.79} & \rankbest{0.793} & \rankbest{0.209} \\[0.5pt]
night\_03 (3843) & \rankbest{27.01} & \rankbest{0.902} & \rankbest{0.148} & 17.96 & 0.513 & 0.472 & 18.56 & 0.727 & 0.478 & \ranksecond{25.55} & \ranksecond{0.876} & \ranksecond{0.168} \\[0.5pt]
night\_04 (1734) & \ranksecond{24.65} & \ranksecond{0.878} & \ranksecond{0.194} & 13.62 & 0.495 & 0.642 & 18.51 & 0.781 & 0.481 & \rankbest{25.77} & \rankbest{0.894} & \rankbest{0.157} \\[0.5pt]
\end{longtable}
\endgroup

\begingroup
\fontsize{9}{10.5}\selectfont
\setlength{\tabcolsep}{3pt}
\renewcommand{\arraystretch}{1.02}
\setlength{\LTleft}{\fill}
\setlength{\LTright}{\fill}
\setlength{\LTcapwidth}{\linewidth}
\begin{longtable}{@{}l*{4}{@{\hspace{10pt}}c}@{}}
\caption{\textbf{Per-sequence geometric reconstruction evaluation.} Depth-L1 is the mean absolute radial-depth error against LiDAR observations (m, $\downarrow$).}
\label{tab:supp-sequence-geometry}\\
\toprule
\shortstack[l]{Sequence\\(image counts)} & Gaussian-LIC2 & ODGS-SLAM & PanoGS-SLAM & \textbf{GOLD-SLAM} \\
\midrule
\endfirsthead
\multicolumn{5}{c}{\tablename~\thetable\ (continued)}\\
\toprule
\shortstack[l]{Sequence\\(image counts)} & Gaussian-LIC2 & ODGS-SLAM & PanoGS-SLAM & \textbf{GOLD-SLAM} \\
\midrule
\endhead
\midrule
\multicolumn{5}{r}{Continued on next page}\\
\endfoot
\bottomrule
\endlastfoot
\multicolumn{5}{@{}l}{\textbf{Hilti-Oxford 2022}}\\*
\addlinespace[2pt]
Exp04 (1257) & \ranksecond{1.997} & 4.542 & 5.462 & \rankbest{1.873} \\[1pt]
Exp14 (740) & \ranksecond{0.426} & 2.278 & 2.501 & \rankbest{0.267} \\[1pt]
Exp18 (1093) & \ranksecond{0.669} & 2.149 & 2.691 & \rankbest{0.450} \\[1pt]
Exp21 (1527) & \ranksecond{1.114} & 9.643 & 11.396 & \rankbest{0.887} \\[1pt]
\midrule
\multicolumn{5}{@{}l}{\textbf{M3DGR}}\\*
\addlinespace[2pt]
Grass01 (4318) & \ranksecond{6.199} & 12.338 & 11.576 & \rankbest{5.992} \\[1pt]
Grass02 (2594) & \ranksecond{7.738} & 11.771 & 11.312 & \rankbest{6.980} \\[1pt]
Varying-illu05 (4000) & \ranksecond{1.594} & 3.447 & 7.513 & \rankbest{0.819} \\[1pt]
Corridor01 (6058) & \ranksecond{1.435} & 1.525 & 3.568 & \rankbest{1.236} \\[1pt]
Corridor02 (4400) & \ranksecond{1.470} & 1.579 & 4.042 & \rankbest{1.206} \\[1pt]
\midrule
\multicolumn{5}{@{}l}{\textbf{M2DGR}}\\*
\addlinespace[2pt]
room\_01 (724) & \ranksecond{0.507} & 1.285 & 1.517 & \rankbest{0.433} \\[1pt]
door\_02 (1256) & \ranksecond{1.334} & 4.574 & 9.879 & \rankbest{1.035} \\[1pt]
hall\_02 (1279) & \ranksecond{1.495} & 5.346 & 10.229 & \rankbest{1.058} \\[1pt]
gate\_03 (2771) & \ranksecond{4.896} & OOM & 25.011 & \rankbest{4.088} \\[1pt]
\midrule
\multicolumn{5}{@{}l}{\textbf{Oxford Spires}}\\*
\addlinespace[2pt]
Keble03 (5601) & \ranksecond{1.249} & OOM & 8.583 & \rankbest{0.357} \\[1pt]
Observatory01 (5632) & \ranksecond{1.681} & 2.795 & 9.786 & \rankbest{0.727} \\[1pt]
ChristChurch03 (6077) & \ranksecond{0.828} & 2.274 & 6.167 & \rankbest{0.449} \\[1pt]
Blenheim01 (7894) & \ranksecond{1.151} & OOM & 8.887 & \rankbest{0.523} \\[1pt]
\midrule
\multicolumn{5}{@{}l}{\textbf{WHOLE}}\\*
\addlinespace[2pt]
outdoor\_01 (2881) & \ranksecond{2.774} & 12.173 & 15.625 & \rankbest{2.128} \\[1pt]
indoor\_01 (1765) & \ranksecond{0.669} & 1.415 & 4.637 & \rankbest{0.434} \\[1pt]
indoor\_02 (1528) & \ranksecond{0.821} & 1.924 & 5.518 & \rankbest{0.452} \\[1pt]
motion\_blur\_01 (1916) & \ranksecond{0.935} & 2.707 & 5.078 & \rankbest{0.587} \\[1pt]
outdoor\_02 (1138) & \ranksecond{1.581} & 3.766 & 11.152 & \rankbest{1.188} \\[1pt]
outdoor\_03 (1663) & \ranksecond{1.557} & 4.569 & 14.136 & \rankbest{1.123} \\[1pt]
motion\_blur\_02 (1547) & \ranksecond{1.918} & 6.689 & 10.792 & \rankbest{1.488} \\[1pt]
motion\_blur\_03 (596) & \ranksecond{2.284} & 4.388 & 10.422 & \rankbest{1.845} \\[1pt]
night\_01 (1602) & \ranksecond{0.585} & 3.965 & 6.172 & \rankbest{0.441} \\[1pt]
night\_02 (2474) & \ranksecond{0.575} & 3.426 & 5.796 & \rankbest{0.417} \\[1pt]
corridor\_01 (3395) & \ranksecond{0.705} & 1.751 & 4.243 & \rankbest{0.393} \\[1pt]
motion\_blur\_04 (1409) & \ranksecond{1.185} & 4.446 & 6.190 & \rankbest{0.767} \\[1pt]
garage\_01 (2900) & \ranksecond{1.470} & 2.611 & 6.965 & \rankbest{0.986} \\[1pt]
garage\_02 (4712) & \ranksecond{1.108} & 2.721 & 6.992 & \rankbest{0.841} \\[1pt]
garage\_03 (2330) & \ranksecond{0.911} & 1.849 & 5.504 & \rankbest{0.692} \\[1pt]
garage\_04 (2810) & \ranksecond{0.783} & 2.454 & 6.423 & \rankbest{0.635} \\[1pt]
outdoor\_04 (4146) & \ranksecond{2.757} & 10.336 & 14.712 & \rankbest{2.378} \\[1pt]
outdoor\_05 (3578) & \ranksecond{3.137} & 10.711 & 15.579 & \rankbest{2.624} \\[1pt]
night\_03 (3843) & \ranksecond{2.324} & 10.697 & 12.087 & \rankbest{2.266} \\[1pt]
night\_04 (1734) & \ranksecond{1.020} & 4.821 & 6.535 & \rankbest{0.721} \\[1pt]
\end{longtable}
\endgroup

{
    \footnotesize
    \setlength{\bibsep}{2pt plus 1pt}
    \setlength{\parskip}{0pt}
    \bibliographystyle{arxiv/refs/ieeenat_fullname}
    \bibliography{arxiv/refs/main}
}

\end{document}